\documentclass[twoside]{article}

\usepackage{linkedin_paper}

\usepackage{booktabs}
\usepackage{colortbl}
\usepackage{pifont}
\usepackage{amsfonts}
\usepackage{amsmath}
\usepackage{amssymb}
\usepackage{mathtools}
\usepackage{amsthm}
\usepackage{nicefrac}
\usepackage{caption}
\usepackage{subcaption}
\usepackage{wrapfig}
\usepackage{makecell}
\usepackage{enumitem}
\usepackage{algorithm}
\usepackage{algpseudocode}
\usepackage[capitalize,noabbrev]{cleveref}

\newlength{\arxivcolw}
\definecolor{darkblue}{rgb}{0, 0, 0.5}
\hypersetup{colorlinks=true, citecolor=darkblue, linkcolor=black, urlcolor=black}

\newcommand{\pp}[1]{\vspace{1.2mm}\noindent\textbf{#1.}}

\newcommand{\sparagraph}[1]{\vspace{0.0mm}\noindent\textbf{#1.}}

\newcommand{\ours}{\mbox{\textsc{ModHarness-RL}}}

\newcommand{\StateN}{\textsc{State}}
\newcommand{\Compression}{\textsc{Compression}}
\newcommand{\Steering}{\textsc{Steering}}
\newcommand{\Replay}{\textsc{Replay}}

\definecolor{compressink}{HTML}{0F766E}
\definecolor{steerink}{HTML}{2563A6}
\definecolor{compressbg}{HTML}{DBEDEA}
\definecolor{steerbg}{HTML}{DEE8F4}
\definecolor{policybg}{HTML}{E9EEF3}
\definecolor{oursrow}{HTML}{EDF2F7}

\newcommand{\frozen}{\,\ding{100}}

\newcommand{\trained}{\,\begin{tikzpicture}[baseline=0pt,x=1em,y=0.88em]
  \fill (0,0) .. controls (0.21,0.04) and (0.25,0.22) .. (0.13,0.42)
        .. controls (0.08,0.56) and (0.10,0.66) .. (0.03,0.80)
        .. controls (0.03,0.62) and (-0.06,0.56) .. (-0.08,0.61)
        .. controls (-0.11,0.44) and (-0.02,0.38) .. (-0.06,0.26)
        .. controls (-0.11,0.34) and (-0.13,0.44) .. (-0.16,0.38)
        .. controls (-0.22,0.26) and (-0.19,0.09) .. (0,0);
\end{tikzpicture}}
\newcommand{\opbox}[2]{%
  \begingroup\setlength{\fboxsep}{1.2pt}\colorbox{#1}{#2}\endgroup%
}
\newcommand{\compress}{\opbox{compressbg}{\texttt{compress}}}
\newcommand{\steer}{\opbox{steerbg}{\texttt{steer}}}

\newcommand{\Dtrain}{\mathcal{D}_{\mathrm{train}}}
\newcommand{\Dhold}{\mathcal{D}_{\mathrm{train}}^{\mathrm{hold}}}

\algrenewcommand{\algorithmiccomment}[1]{\hfill\textcolor{black!55}{\footnotesize\# #1}}

\usepackage{tcolorbox}
\tcbuselibrary{breakable,skins}
\def\promptink{black}
\newtcolorbox{promptbox}[2]{breakable, enhanced, colback=black!1.5, colframe=#2ink, coltitle=white,
  colbacktitle=#2ink, fonttitle=\bfseries\small, title={#1}, boxrule=0.6pt, arc=2pt,
  left=8pt, right=8pt, top=6pt, bottom=6pt, toptitle=2pt, bottomtitle=2pt,
  fontupper=\footnotesize\ttfamily, before upper={\setlength{\parskip}{4pt}\def\promptink{#2ink}},
  before skip=10pt, after skip=10pt}
\newcommand{\slot}[1]{{\color{\promptink!80}$\langle$\textit{#1}$\rangle$}}
\newcommand{\prompthead}[1]{\par\vspace{3pt}\noindent{\color{\promptink}\bfseries #1}\par\vspace{-2pt}}

\newif\ifonecolumn
\newcommand{\wraptab}[1]{{\renewenvironment{table}[1][]{}{}\setlength{\arxivcolw}{\linewidth}#1}}
\newcommand{\circnum}[1]{\tikz[baseline=(c.base)]\node[circle,draw,line width=0.5pt,inner sep=0pt,minimum size=1.05em,font=\footnotesize](c){#1};}

\onecolumntrue
\newcommand{\fAlfLimGain}{21}
\newcommand{\fLimCtxWithin}{3}
\newcommand{\fCsLimGain}{8}
\newcommand{\fAlfSufSave}{74}

\newcommand{\fCsSufSave}{10}

\newcommand{\fCsCompLo}{5}
\newcommand{\fCsCompHi}{12}

\newcommand{\fCsOursAvg}{19}
\newcommand{\fCsOursLo}{14}
\newcommand{\fCsOursHi}{24}
\newcommand{\fOtherUp}{seven}
\newcommand{\fOtherCells}{eight}
\newcommand{\fBenchCsSteer}{15}
\newcommand{\fBenchCsComp}{6}
\newcommand{\fBenchAlfLimComp}{22}
\newcommand{\fBenchAlfLimSteer}{16}

\newcommand{\fCmpCalls}{100}

\newcommand{\fCmpLat}{1,000}
\newcommand{\fCmpBehind}{5}
\newcommand{\fCmpBehindN}{two}
\newcommand{\fCmpAhead}{2}
\newcommand{\fCmpCells}{4}

\newcommand{\fKvReuseLo}{58}
\newcommand{\fKvReuseHi}{72}
\newcommand{\fKvBaseLo}{64}
\newcommand{\fKvBaseHi}{95}
\newcommand{\fKvCells}{4}
\newcommand{\fKvCheaper}{3}
\newcommand{\fKvBestSave}{35}

\newcommand{\fSqlFireLo}{30}
\newcommand{\fSqlFireHi}{42}

\title{\ours: Recasting Harnesses as Learnable\\ and Evolving Programs for Agentic RL}

\author{%
\small{Zelin He$^{1,*,\dagger}$, Siyu Zhu$^{2,*}$, Jacob Helwig$^{2,*}$, Yujian Liu$^{3,\dagger}$,\\[0.45em]
Li An$^{3,\dagger}$, Rohit Jain$^{2}$, Shiyu Chang$^{2,3}$, Alborz Geramifard$^{2}$}%
  \\[0.8em]
  {\fontsize{10pt}{11pt}\selectfont
  \href{https://zlhe0.github.io/ModHarness-RL/}{\raisebox{-0.15em}{\includegraphics[height=1em]{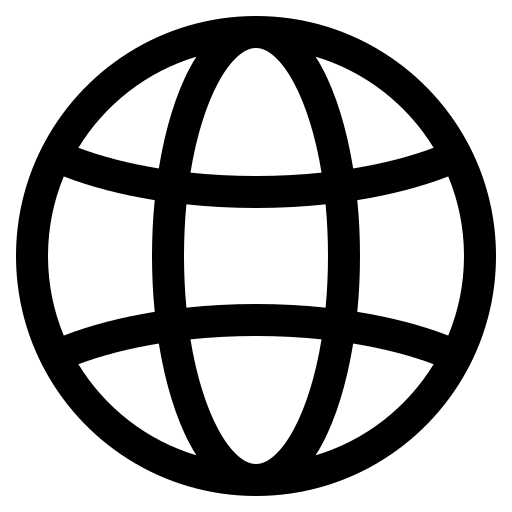}}\,Website}}
}

\begin{document}

\maketitle
\begingroup
\renewcommand{\thefootnote}{}
\footnotetext{\hspace{-1.8em}\begin{tabular}{@{}l@{}}
$^{*}$ Equal contribution.\\[0.15em]
$^{\dagger}$ Work done during an internship at LinkedIn.\\[0.15em]
$^{1}$ The Pennsylvania State University \quad $^{2}$ LinkedIn \quad $^{3}$ University of California, Santa Barbara
\end{tabular}}
\endgroup
\vspace{-1.0em}

\begin{abstract}
Agentic reinforcement learning (RL) increasingly relies on a harness, the scaffold around
the model that decides what it sees of its own interaction and what guidance it receives.
Most existing approaches train under a hand-written harness that stays fixed throughout.
However, an effective harness depends on the domain, the data and the model, and keeps
changing as the model trains. We introduce \ours, which recasts the harness as programs that are
learned from data and evolve as the model trains with RL. \ours\ records the interaction as
structured \StateN, a set of typed blocks, and two core components of the harness are programs
over these blocks, a \Compression\ program that assembles the accumulated history into the model's context and a \Steering\ program that supplies guidance when
behavior deviates. During RL training, candidate programs are proposed from recent
trajectories, verified by execution, and replayed over past trajectories, so the harness
tracks how the model's behavior shifts. We study CodeSearch in depth and further analyze
ALFWorld and BIRD SQL, with models from 1.5B to 8B. The proposed \Compression\ and \Steering\
mechanisms improve held-out performance over the fixed-harness baseline in most settings.
These results suggest that a harness for
agentic RL can be learned from the task's data and evolved with the model.
\end{abstract}

\par\vspace{0.4em}\noindent\begin{minipage}{\textwidth}\centering
\includegraphics[width=\textwidth,height=2.28in,keepaspectratio]{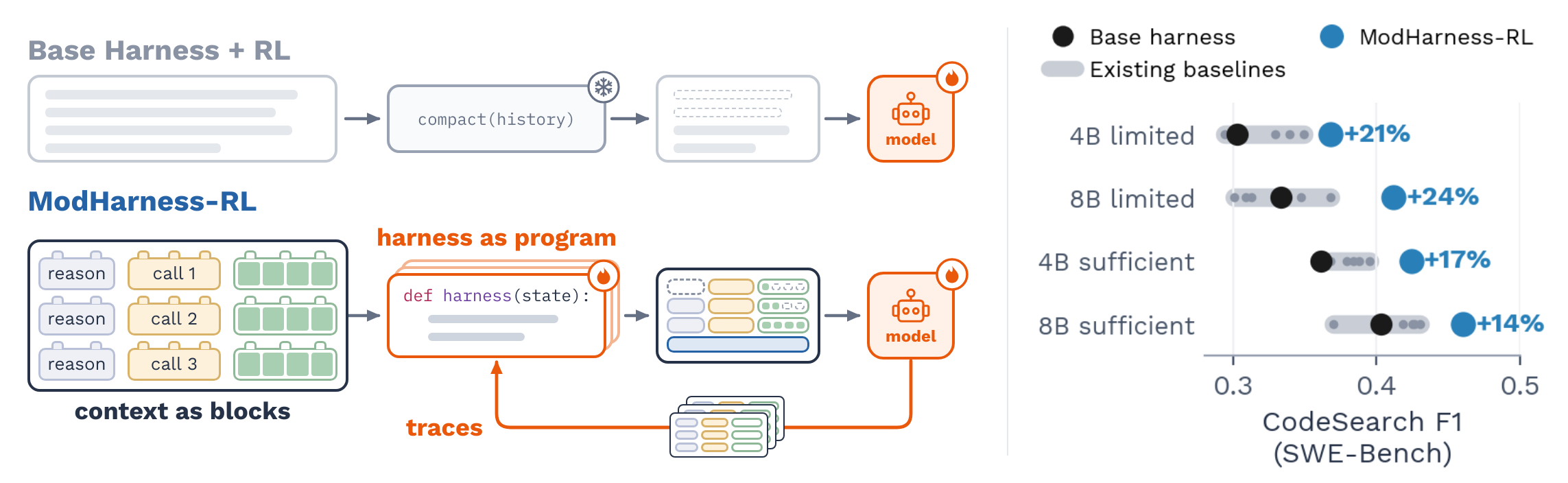}
\captionof{figure}{Left: Base Harness + RL (top) and \ours\ (bottom). \ours\ exploits the modularity of an agent's interaction to learn the harness from its trajectories and evolve it with the model. Right: Performance improvement of \ours\ on CodeSearch over SWE-Bench Lite, Verified and Pro, with a context budget that runs out mid-episode (limited) or is rarely reached (sufficient).}\label{fig:headline}
\end{minipage}\par\vspace{0.2em}

\section{Introduction}

Large language models now act as autonomous agents that solve tasks through
multi-turn interaction with external environments~\citep{yao2023react,liu2024agentbench}.
Agentic reinforcement learning (RL), which trains these agents directly from
verifiable environment rewards, has emerged as a promising paradigm
~\citep{jin2025searchr1,wei2025webagentr1,da2025agentrlvr}. A harness, the engineered
layer between the environment and the model, governs what the model reads and how its
behavior is guided.
One line of work
compacts the growing transcript of the interaction through memory, summarization, and pruning
~\citep{packer2024memgpt,yu2025resum,rlm2025,bashir2026context1,shao2026clm}, while another
guides behavior through added instructions, interfaces, and guardrails
~\citep{yang2024sweagent,xia2024agentlessdemystifyingllmbasedsoftware}. Agentic RL
has begun to train models under such harnesses, but the harness is typically
hardcoded for a specific task and held fixed throughout training
~\citep{jin2025searchr1,qian2025toolrl,qin2023toolllm}.
This leaves two challenges.
First, an effective harness must be tailored to the task, the data, the model, and the
context budget, and tuning one by hand for every such configuration does not scale, so it
should be derived from data instead.
Second, the effective harness keeps changing as the model trains. RL resolves some
failures on its own, so new harness components should be added only for the failures that persist.
This leads to a natural question. \textit{How can a harness be made
optimizable from data, and then optimized with the model training?}

Recent efforts have begun to adapt the harness during RL, by decaying fixed rubric
scaffolds on a preset schedule~\citep{qiu2025ruscarl}, alternating task-specific scaffold
rewrites with weight updates~\citep{hebbar2026sia}, or editing planning, action and memory
modules from observed faults~\citep{chen2026harnessforge}. Yet their update spaces are fixed
in advance or tied to a task-specific scaffold, so adaptation reduces to selection or
revision within an engineered space. None provides a general way to derive harness
operations from the typed states of the interaction and optimize them alongside the model.

In this paper we propose \ours, a method that exploits the modularity of an agent's interaction to learn the harness from its trajectories and evolve it with the model. It is built around two questions.

\begin{itemize}[leftmargin=1.2em,itemsep=2pt,topsep=3pt]
\item \textbf{How should a harness be represented to be learned and evolved?} To learn a harness from data, we first need a representation of it that defines what can be searched and evolved, and we need to narrow which of its components to search. \ours\ represents the harness as programs over a structured
\StateN, which records the interaction as typed blocks such as each turn's reasoning, tool call
and tool result. These blocks are the building blocks from which \ours\ assembles the context and control. We focus on two core components, both written as programs over them. A \Compression\ program reassembles the blocks into organized model context under a
fixed budget, and a \Steering\ program reads them to detect when behavior deviates and appends guidance.
Programs over these blocks have three properties that suit learning and evolving. They are
(1) learnable, since an LLM can write them directly from training trajectories and their corresponding rewards; (2) computationally inexpensive, allowing them to be run on every turn of every trajectory without
calling another model, so \Compression\ can optimize the context at scale; and
(3) testable, allowing each candidate harness to be replayed over stored trajectories to evaluate its fitness before it is accepted or rejected.

\begin{figure*}[t]
	\centering
	\includegraphics[width=\linewidth]{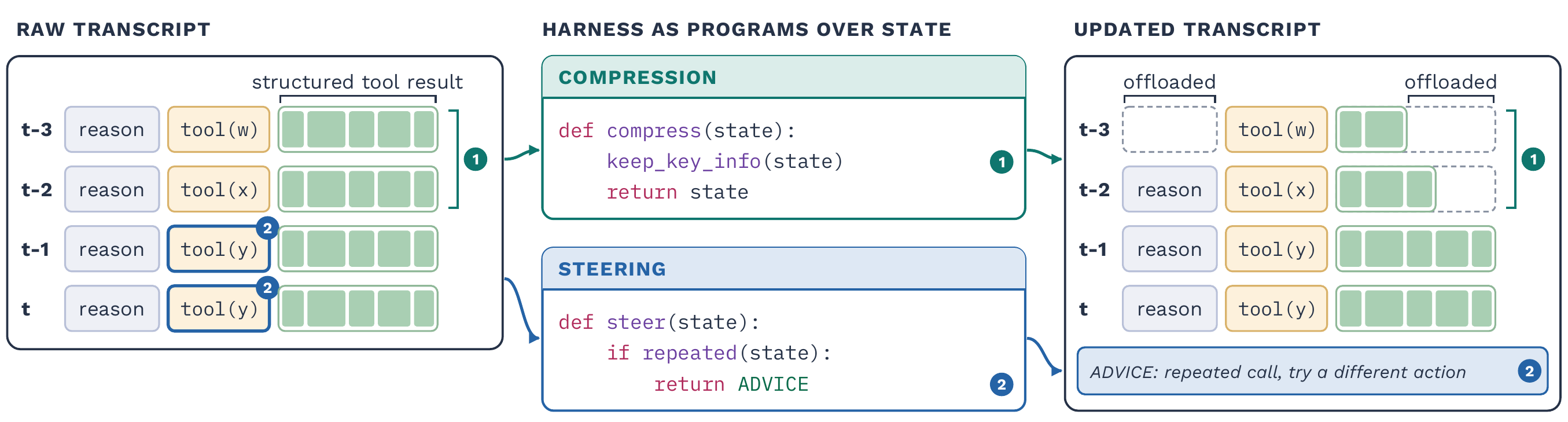}
	\caption{\textbf{The harness as programs over \StateN.} Compression and Steering are
	programs that read the blocks of \StateN. Compression returns a new view of \StateN\ that re-renders earlier turns, keeping
	their key information and offloading what is no longer needed. Steering checks the trace
	for specific conditions and, when one holds, returns a message that is appended to steer the model's
	behavior.}
	\label{fig:harness_as_code}
\end{figure*}

\item \textbf{How can the harness be learned and evolved with the model?} As training progresses, model behaviors change. Failure modes from previous model versions may become less frequent, while new ones may emerge. \ours{} therefore uses an
end-to-end pipeline that updates the harness from online data. It summarizes experiences from training
trajectories, proposes and verifies candidate programs, and validates them on held-out training tasks.
To evolve the harness with the model, we alternate this pipeline with RL updates of the model
weights. \textit{The key insight is that the harness should guide only what the model cannot
learn on its own.} Therefore, we propose a \Replay\ mechanism, which reruns each candidate
\Steering\ program over trajectories from past training steps and installs it only for the
failures that persist as the model trains.
\end{itemize}

We train models from 1.5B to 8B with GRPO in three environments (CodeSearch, ALFWorld and BIRD SQL),
each under different context budgets. On CodeSearch, \ours\ improves over the fixed-harness
baseline by \fCsOursLo\ to \fCsOursHi\% in every setting and scores highest against memory,
skill, prompt and harness evolution baselines, by \fCsCompLo\ to \fCsCompHi\% over the
strongest of them. On ALFWorld and BIRD SQL, each program alone improves over the base
harness in \fOtherUp\ of \fOtherCells\ settings. We also investigate a series of research questions, from
the score and token use of \Compression\ and its comparison with frontier
model-based compaction, to an ablation of each \Steering\ component.

\section{Preliminaries}
\label{sec:prelim}

\pp{Multi-Turn Agents}
An LLM agent with policy $\pi_\theta$ solves a task $q \sim \mathcal{D}$ by
interacting with an environment over multiple turns through a set of tools
$\mathcal{T}$. At turn $t$ the agent reads a context $c_t$, samples an action
$a_t \sim \pi_\theta(\cdot \mid c_t)$, its full response for the turn ending in a tool call, and receives an observation
$o_t = \mathcal{T}(a_t)$, continuing until a terminal turn $T$. The harness builds the context $c_t$, including any
guidance the agent receives, from the task and the interaction so far.

\pp{Agentic Reinforcement Learning}
The agent is trained to maximize the expected reward of its trajectory
$\tau = (q, a_0, o_0, \ldots, a_T)$, that is
$\max_\theta \mathbb{E}_{q \sim \mathcal{D},\, \tau \sim \pi_\theta}[r(\tau)]$, where $r(\tau)$ is
the reward that the environment assigns to the trajectory. In this paper, we optimize this objective with GRPO~(\citealp{shao2024deepseekmath}; \Cref{app:grpo}).

\section{Learning and Evolving the Harness as Programs in Agentic RL}
\label{sec:method}

We introduce \ours, which learns the harness from data and evolves it with the model as the model trains with RL. It represents the harness as programs over the typed blocks of a
structured \StateN\ for two core components (\Cref{sec:harness_code}), and learns and evolves
these programs with the model.

\subsection{Representing the Harness as Programs over State}
\label{sec:harness_code}
As the agent interacts with the environment, we record each turn as an action and
observation pair $u_i=(a_i,o_i)$ and take the \StateN\ to be the task together with the turns
so far, $s_t = (q, u_0, \ldots, u_{t-1})$. We store $s_t$ as a typed record, like an instance
of a Python class, where each turn has attributes for the reasoning the model wrote
(\texttt{.thought}), the tool call (\texttt{.name}, \texttt{.args}), and the result
(\texttt{.exec\_output}, or retrieved documents with their own typed metadata), plus any
guidance shown on that turn (\Cref{fig:harness_as_code}, left). A basic harness renders this record directly
as one growing transcript, appending every turn,
\begin{equation}\label{eq:appendOnly}
c_t^{\mathrm{flat}}=\operatorname{flatten}(s_t)=q\oplus u_0\oplus\cdots\oplus u_{t-1},
\end{equation}
where $\oplus$ denotes concatenation. This append-only context leaves two problems, and we
address each with a program over $s_t$.

\begin{description}[leftmargin=*, itemsep=2pt, topsep=3pt]
\item[\compress] As the interaction accumulates, a high volume of irrelevant tool outputs may obfuscate the small fraction of decision-relevant evidence~\citep{liu2024lostmiddle}. \compress\ reads the blocks of $s_t$ and returns a new view
of the \StateN\ that keeps the key information and fits the budget. It re-renders earlier
turns and offloads what is no longer needed (\Cref{fig:harness_as_code}, \circnum{1}).

\item[\steer] Without any intervention, models sometimes show recurring mistakes~\citep{yang2024sweagent}. \steer\ is a set of triggers, each
paired with a short piece of guidance. A trigger is a small program over $s_t$ that fires on
one detectable failure mode, such as a repeated tool call, and \steer\ returns its guidance as
a block only on the turns where it fires (\Cref{fig:harness_as_code}, \circnum{2}).
\end{description}

\noindent Together the two programs form the harness we learn, $\mathcal{H} = (\compress,\ \steer)$.
The rest of the harness, such as the tool set and the action format, remains fixed as each
environment defines it. In place of
$c_t^{\mathrm{flat}}$, the harness renders the new view of the \StateN\ that the two programs produce,
\begin{equation}
\label{eq:policy_view}
c_t = \operatorname{flatten}\bigl(\compress(s_t) \oplus \steer(s_t)\bigr),
\end{equation}
where the guidance is appended after the compressed turns and is empty on the turns when no
trigger fires. Both programs read the full record $s_t$ at every turn, so what one turn
offloads stays available for building later contexts.
 Executing $\mathcal{H}$ calls no model and costs no model tokens, so it
runs on every turn of every trajectory at scale.
See \Cref{app:programs} for concrete examples.

\newcommand{\oursalgorithm}[1]{%
\begin{algorithm}[#1]
\caption{\ours}
\label{alg:ours}
\small
\begin{algorithmic}[1]
\Require Base policy $\pi_0$; rounds $K$ ; Training tasks $\Dtrain$; held-out training tasks $\Dhold$

\Statex Compression optimization
\State $\mathcal{R}_0 \gets \Call{Rollout}{\pi_0, \Dtrain}$
\State $\mathcal{P} \gets \Call{ProposeCompress}{\mathcal{R}_0}$
\State $\mathcal{H}_0 \gets \Call{Gate}{\pi_0, \varnothing, \mathcal{P}, \Dhold}$

\Statex \hrulefill
\Statex Alternate policy and steering optimization
\For{$k=1,\ldots,K$}
    \Statex \hspace{\algorithmicindent} %
    \State $(\pi_k,\mathcal{R}_k) \gets \Call{UpdatePolicy}{\pi_{k-1},\mathcal{H}_{k-1},\Dtrain}$
    \Statex \hspace{\algorithmicindent} %
    \State $\mathcal{P} \gets \Call{ProposeSteer}{\mathcal{R}_k}$
    \State $\mathcal{E} \gets \Call{Replay}{\mathcal{P},\mathcal{H}_{k-1},\mathcal{R}_{0:k}}$
    \State $\mathcal{P} \gets \Call{Admit}{\mathcal{P},\mathcal{H}_{k-1},\mathcal{E}}$
    \State $\mathcal{H}_k \gets \Call{Gate}{\pi_k,\mathcal{H}_{k-1},\mathcal{P},\Dhold}$
\EndFor
\State \Return $(\pi_K,\mathcal{H}_K)$
\end{algorithmic}
\end{algorithm}
}

\subsection{Evolving the Harness with the Policy}
\label{sec:coevolution}
\begin{figure*}[t]
	\centering
	\includegraphics[width=\linewidth]{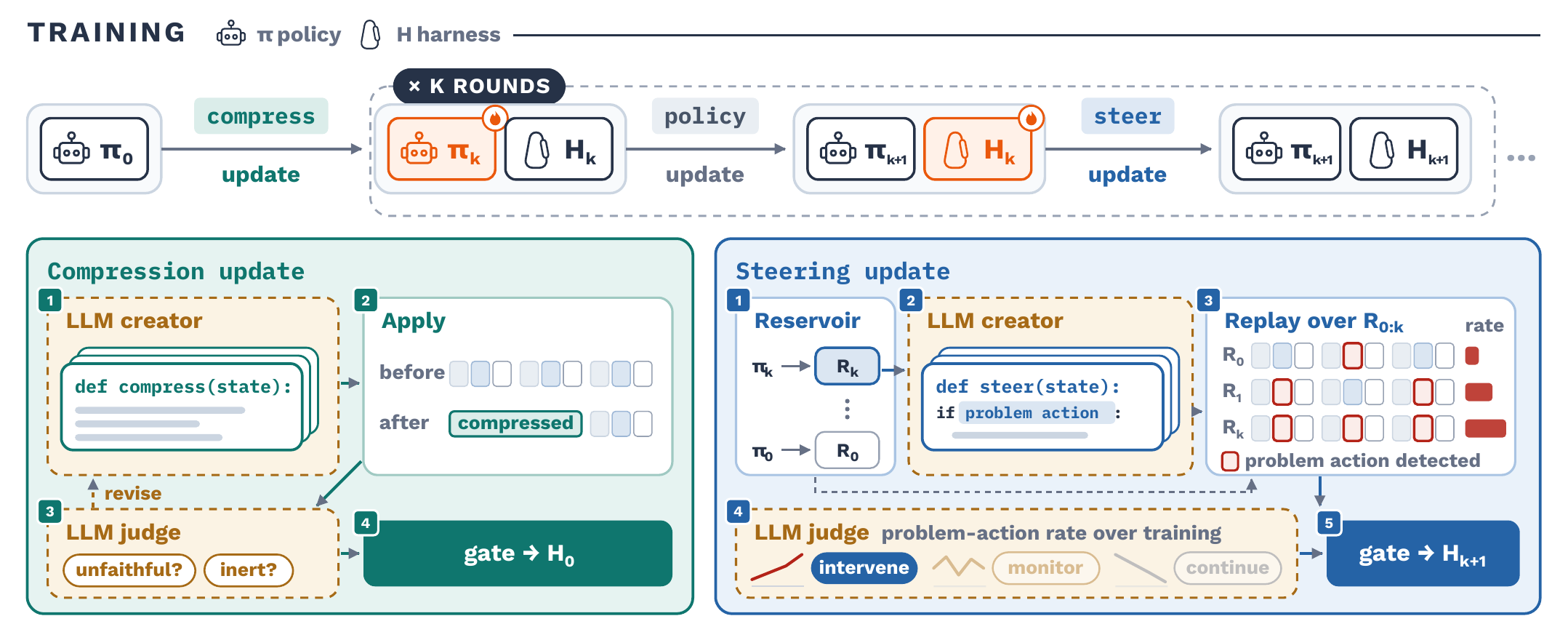}
    \caption{\textbf{Learning and evolving the harness with the policy.} \compress\ is optimized against the base policy, and policy updates under the latest harness are then alternated with \steer\ updates. Both optimizations propose
	programs from the reservoir, verify them by execution, and install a winner only
	through a held-out gate.}
	\label{fig:staged_optimization}
\end{figure*}

With the harness represented as programs, we can build these programs from the policy's own
rollouts, so that they are tailored to both the task and the policy, and keep evolving them
over training (\Cref{fig:staged_optimization}, \Cref{alg:ours}). 

\ifonecolumn\else\oursalgorithm{t}\fi

\subsubsection{Optimizing the Compression Harness}
\label{sec:compression}

Verbose tool outputs waste context tokens and distract the model from decision-relevant evidence, limiting performance. The compression optimization performs a data-driven search for a program that retains only the spans of tool outputs driving task performance. We fit \compress\ once before training and hold it fixed,
so that the context the policy trains on stays stable. While we also explored refitting
\compress\ as the policy trains, we found that holding it fixed scores higher
(\Cref{tab:ablation}).

\ifonecolumn\begin{wrapfigure}[19]{r}{0.5\textwidth}\vspace{-14pt}\begin{minipage}{\linewidth}\small\oursalgorithm{H}\end{minipage}\end{wrapfigure}\fi
\pp{\textsc{Rollout}}
We roll out the base policy $\pi_0$ on the training tasks $\Dtrain$ under the
append-only context of \Cref{eq:appendOnly}, keeping each trajectory with its reward and
grouping them by task. The result seeds the reservoir
$\mathcal{R}_0 = \textsc{Rollout}(\pi_0, \Dtrain)$.

\pp{\textsc{ProposeCompress}}
The rollouts are fed to an LLM creator. The creator reads these rollouts, along with how their contexts filled up, and writes several candidate compression programs. Each candidate
is then refined in a loop. The creator compiles and repairs each candidate until it runs, applies it to stored
episodes from $\mathcal{R}_0$, and computes two measurements. One checks whether the program
is \emph{unfaithful}, removing task-relevant information, and the other checks whether it
is \emph{inert}, leaving the context largely unchanged. The creator accepts the candidate or
revises it, and the last version of each candidate forms the proposal set $\mathcal{P}$.

\pp{\textsc{Gate}}
Each program in $\mathcal{P}$ gets one trial on the held-out training tasks $\Dhold$ under the
frozen base policy, with the same agent loop, tools and context limit that training will
use. The best becomes $\mathcal{H}_0$ and is held fixed for the rest of training.

\subsubsection{Evolving the Steering Harness with the Policy}
\label{sec:steering}

Steering optimization guides the policy only on the failures it cannot learn to fix during
training. These failures change as the policy trains, so \steer\ is updated every round.

\pp{\textsc{UpdatePolicy}}
 We train $\pi_{k-1}$ with GRPO on $\Dtrain$
under the incumbent harness $\mathcal{H}_{k-1}$, which yields the updated policy
$\pi_k$ together with the rollouts $\mathcal{R}_k$ it produced along the way, and those rollouts join the reservoir. 

\pp{\textsc{ProposeSteer}}
$\mathcal{R}_k$ records what the current policy can and cannot do, so the creator reads
its mixed-outcome groups, the tasks on which some rollouts succeed and others fail, and
proposes failure hypotheses from them. Each hypothesis carries a trigger, a small program
over $s_t$, together with the corrective insight its guidance will be written from.
The proposed triggers form the proposal set $\mathcal{P}$.

\pp{\textsc{Replay}}
The newest rollouts show what is failing now, but not whether the failure will persist. It
may be an incidental error that the policy learns to avoid with further training, or a
stable failure that it will not fix on its own. Because each trigger is a program, we can
classify the failure type retrospectively. We execute every candidate hypothesis over all stored
rollouts $\mathcal{R}_{0:k}$ and record how often its failure occurs at each
training step. Replay needs no model call, so it stays computationally inexpensive however large the reservoir
grows. The resulting curves show the trend of each failure over training, whether it
is falling, holding steady or rising, and the curves of all triggers together give the
replay evidence $\mathcal{E}$.

\pp{\textsc{Admit}}
A judge reads each trigger's replay evidence together with its guidance and the installed
set, and returns one of three verdicts: (1) \emph{intervene} when the failure holds steady
or rises, installing the trigger and its guidance; (2) \emph{monitor} when it looks real but
has not settled, keeping the trigger for another round without showing guidance; or
(3) \emph{continue} when it keeps falling. 

\pp{\textsc{Gate}}
The admitted triggers are gated together, since guidance interacts, and accepted as
$\mathcal{H}_k$ only if they score strictly above $\mathcal{H}_{k-1}$ on the held-out training tasks
$\Dhold$ under $\pi_k$.

\section{Experiments}
\label{sec:experiments}

In this section, we evaluate the performance and efficiency of \ours\ on three agentic environments (Section~\ref{sec:main_results}), and then study its two harness modules through three research questions (Section~\ref{sec:analysis}).

\begin{figure*}[t]
\centering
\includegraphics[width=\textwidth]{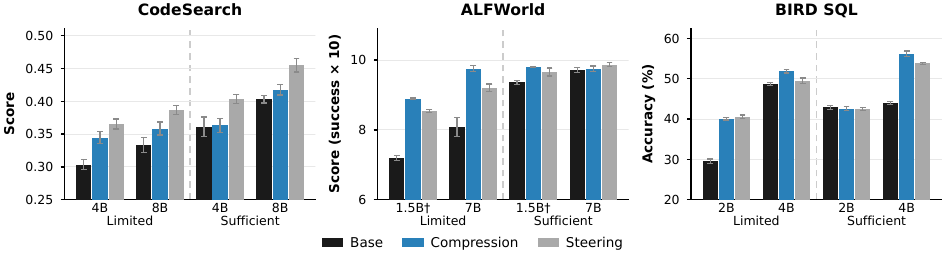}
\caption{\textbf{\Compression\ and \Steering\ each applied to the base harness on three
benchmarks.} Models are Qwen3 on
CodeSearch, Qwen2.5 on ALFWorld and Qwen3.5 on BIRD SQL, and the tick gives the scale. Whiskers are $\pm$1 SD over repeated evaluations of each trained checkpoint. \textsuperscript{\dag}SFT warm start.}
\label{fig:benchmarks}
\end{figure*}

\subsection{Setup}
\ifonecolumn\begin{table}[t]
\centering
\footnotesize
\setlength{\tabcolsep}{2.6pt}
\begin{tabular*}{\textwidth}{@{\extracolsep{\fill}}l*{10}{c}@{}}
\toprule
& \multicolumn{5}{c}{\itshape Context-limited} & \multicolumn{5}{c}{\itshape Context-sufficient} \\
\cmidrule(lr){2-6}\cmidrule(l){7-11}
Method & Lite & Verified & Pro & Overall & $\Delta$ & Lite & Verified & Pro & Overall & $\Delta$ \\
\midrule
\multicolumn{11}{@{}l}{\bfseries Qwen3-4B} \\[1pt]
\quad Zero-shot\frozen & 0.172 & 0.139 & 0.113 & 0.142 & \textbf{$-$53\%} & 0.138 & 0.111 & 0.105 & 0.119 & \textbf{$-$67\%} \\
\quad Base harness\trained & 0.349 & 0.328 & 0.224 & 0.303 & --- & 0.385 & 0.351 & 0.346 & 0.361 & --- \\
\quad DeepAgents\trained & 0.358 & 0.334 & 0.292 & 0.329 & \textbf{$+$9\%} & 0.407 & 0.380 & 0.310 & 0.368 & \textbf{$+$2\%} \\
\quad INSPO\trained & 0.323 & 0.317 & 0.253 & 0.299 & \textbf{$-$1\%} & 0.400 & 0.408 & 0.322 & 0.379 & \textbf{$+$5\%} \\
\quad EvolveR\trained & 0.327 & 0.327 & 0.219 & 0.294 & \textbf{$-$3\%} & 0.406 & 0.416 & 0.322 & 0.384 & \textbf{$+$6\%} \\
\quad HarnessForge\trained & 0.372 & 0.351 & 0.289 & 0.339 & \textbf{$+$12\%} & 0.407 & 0.394 & 0.358 & 0.388 & \textbf{$+$7\%} \\
\quad SkillRL\trained & 0.359 & 0.374 & \textbf{0.308} & 0.349 & \textbf{$+$15\%} & 0.438 & 0.400 & 0.341 & 0.395 & \textbf{$+$9\%} \\
\rowcolor{oursrow}\quad + \Compression\trained & 0.393 & 0.356 & 0.277 & 0.345 & \textbf{$+$14\%} & 0.374 & 0.374 & 0.338 & 0.363 & \textbf{$+$1\%} \\
\rowcolor{oursrow}\quad + \Steering\trained & \textbf{0.426} & 0.361 & 0.302 & 0.365 & \textbf{$+$20\%} & 0.439 & 0.410 & 0.356 & 0.403 & \textbf{$+$12\%} \\
\rowcolor{oursrow}\quad \ours\trained & 0.422 & \textbf{0.388} & 0.282 & \textbf{0.368} & \textbf{$+$21\%} & \textbf{0.454} & \textbf{0.432} & \textbf{0.381} & \textbf{0.424} & \textbf{$+$17\%} \\
\addlinespace[2pt]
\multicolumn{11}{@{}l}{\bfseries Qwen3-8B} \\[1pt]
\quad Zero-shot\frozen & 0.090 & 0.084 & 0.095 & 0.089 & \textbf{$-$73\%} & 0.098 & 0.097 & 0.059 & 0.086 & \textbf{$-$79\%} \\
\quad Base harness\trained & 0.363 & 0.357 & 0.272 & 0.333 & --- & 0.431 & 0.419 & 0.352 & 0.403 & --- \\
\quad DeepAgents\trained & 0.433 & 0.371 & 0.289 & 0.368 & \textbf{$+$10\%} & 0.465 & 0.456 & 0.361 & 0.430 & \textbf{$+$7\%} \\
\quad INSPO\trained & 0.331 & 0.358 & 0.241 & 0.313 & \textbf{$-$6\%} & 0.477 & 0.441 & 0.347 & 0.425 & \textbf{$+$5\%} \\
\quad EvolveR\trained & 0.381 & 0.374 & 0.277 & 0.347 & \textbf{$+$4\%} & 0.443 & 0.450 & 0.353 & 0.418 & \textbf{$+$4\%} \\
\quad HarnessForge\trained & 0.350 & 0.348 & 0.192 & 0.301 & \textbf{$-$10\%} & 0.430 & 0.395 & 0.274 & 0.370 & \textbf{$-$8\%} \\
\quad SkillRL\trained & 0.338 & 0.308 & 0.275 & 0.309 & \textbf{$-$7\%} & 0.478 & 0.428 & 0.362 & 0.426 & \textbf{$+$6\%} \\
\rowcolor{oursrow}\quad + \Compression\trained & 0.392 & 0.356 & 0.323 & 0.359 & \textbf{$+$8\%} & 0.453 & 0.461 & 0.329 & 0.418 & \textbf{$+$4\%} \\
\rowcolor{oursrow}\quad + \Steering\trained & \textbf{0.453} & 0.397 & 0.298 & 0.386 & \textbf{$+$16\%} & 0.483 & \textbf{0.487} & 0.385 & 0.455 & \textbf{$+$13\%} \\
\rowcolor{oursrow}\quad \ours\trained & 0.441 & \textbf{0.437} & \textbf{0.351} & \textbf{0.412} & \textbf{$+$24\%} & \textbf{0.492} & 0.473 & \textbf{0.409} & \textbf{0.460} & \textbf{$+$14\%} \\
\addlinespace[2pt]
\multicolumn{11}{@{}l}{\bfseries GPT-5-mini} \\[1pt]
\quad Zero-shot\frozen & 0.113 & 0.108 & 0.055 & 0.094 & --- & 0.404 & 0.389 & 0.298 & 0.366 & --- \\
\bottomrule
\end{tabular*}
\caption{\textbf{CodeSearch validation, F1 averaged over the three
granularities.} $\Delta$ is the relative change in \emph{Overall} over the base
harness of the same model, and bold is best in a column within a model. Shaded rows are
ours. \trained\ marks an RL-trained model and \frozen\ one that was not fine-tuned.}
\label{tab:cs_main}
\end{table}
\else\begin{table}[t]
\centering
\scriptsize
\setlength{\tabcolsep}{2.6pt}
\begin{tabular*}{\arxivcolw}{@{\extracolsep{\fill}}l*{4}{c}c@{}}
\toprule
Method & Lite & Verified & Pro & Overall & $\Delta$ \\
\midrule
\multicolumn{6}{@{}l}{\itshape Context-limited} \\[1pt]
\addlinespace[1pt]
\multicolumn{6}{@{}l}{\bfseries Qwen3-4B} \\[1pt]
\quad Zero-shot\frozen & 0.172 & 0.139 & 0.113 & 0.142 & \textbf{$-$53\%} \\
\quad Base harness\trained & 0.349 & 0.328 & 0.224 & 0.303 & --- \\
\quad DeepAgents\trained & 0.358 & 0.334 & 0.292 & 0.329 & \textbf{$+$9\%} \\
\quad INSPO\trained & 0.323 & 0.317 & 0.253 & 0.299 & \textbf{$-$1\%} \\
\quad EvolveR\trained & 0.327 & 0.327 & 0.219 & 0.294 & \textbf{$-$3\%} \\
\quad HarnessForge\trained & 0.372 & 0.351 & 0.289 & 0.339 & \textbf{$+$12\%} \\
\quad SkillRL\trained & 0.359 & 0.374 & \textbf{0.308} & 0.349 & \textbf{$+$15\%} \\
\rowcolor{oursrow}\quad + \Compression\trained & 0.393 & 0.356 & 0.277 & 0.345 & \textbf{$+$14\%} \\
\rowcolor{oursrow}\quad + \Steering\trained & \textbf{0.426} & 0.361 & 0.302 & 0.365 & \textbf{$+$20\%} \\
\rowcolor{oursrow}\quad \ours\trained & 0.422 & \textbf{0.388} & 0.282 & \textbf{0.368} & \textbf{$+$21\%} \\
\addlinespace[1pt]
\multicolumn{6}{@{}l}{\bfseries Qwen3-8B} \\[1pt]
\quad Zero-shot\frozen & 0.090 & 0.084 & 0.095 & 0.089 & \textbf{$-$73\%} \\
\quad Base harness\trained & 0.363 & 0.357 & 0.272 & 0.333 & --- \\
\quad DeepAgents\trained & 0.433 & 0.371 & 0.289 & 0.368 & \textbf{$+$10\%} \\
\quad INSPO\trained & 0.331 & 0.358 & 0.241 & 0.313 & \textbf{$-$6\%} \\
\quad EvolveR\trained & 0.381 & 0.374 & 0.277 & 0.347 & \textbf{$+$4\%} \\
\quad HarnessForge\trained & 0.350 & 0.348 & 0.192 & 0.301 & \textbf{$-$10\%} \\
\quad SkillRL\trained & 0.338 & 0.308 & 0.275 & 0.309 & \textbf{$-$7\%} \\
\rowcolor{oursrow}\quad + \Compression\trained & 0.392 & 0.356 & 0.323 & 0.359 & \textbf{$+$8\%} \\
\rowcolor{oursrow}\quad + \Steering\trained & \textbf{0.453} & 0.397 & 0.298 & 0.386 & \textbf{$+$16\%} \\
\rowcolor{oursrow}\quad \ours\trained & 0.441 & \textbf{0.437} & \textbf{0.351} & \textbf{0.412} & \textbf{$+$24\%} \\
\addlinespace[1pt]
\multicolumn{6}{@{}l}{\bfseries GPT-5-mini} \\[1pt]
\quad Zero-shot\frozen & 0.113 & 0.108 & 0.055 & 0.094 & --- \\
\midrule
\multicolumn{6}{@{}l}{\itshape Context-sufficient} \\[1pt]
\addlinespace[1pt]
\multicolumn{6}{@{}l}{\bfseries Qwen3-4B} \\[1pt]
\quad Zero-shot\frozen & 0.138 & 0.111 & 0.105 & 0.119 & \textbf{$-$67\%} \\
\quad Base harness\trained & 0.385 & 0.351 & 0.346 & 0.361 & --- \\
\quad DeepAgents\trained & 0.407 & 0.380 & 0.310 & 0.368 & \textbf{$+$2\%} \\
\quad INSPO\trained & 0.400 & 0.408 & 0.322 & 0.379 & \textbf{$+$5\%} \\
\quad EvolveR\trained & 0.406 & 0.416 & 0.322 & 0.384 & \textbf{$+$6\%} \\
\quad HarnessForge\trained & 0.407 & 0.394 & 0.358 & 0.388 & \textbf{$+$7\%} \\
\quad SkillRL\trained & 0.438 & 0.400 & 0.341 & 0.395 & \textbf{$+$9\%} \\
\rowcolor{oursrow}\quad + \Compression\trained & 0.374 & 0.374 & 0.338 & 0.363 & \textbf{$+$1\%} \\
\rowcolor{oursrow}\quad + \Steering\trained & 0.439 & 0.410 & 0.356 & 0.403 & \textbf{$+$12\%} \\
\rowcolor{oursrow}\quad \ours\trained & \textbf{0.454} & \textbf{0.432} & \textbf{0.381} & \textbf{0.424} & \textbf{$+$17\%} \\
\addlinespace[1pt]
\multicolumn{6}{@{}l}{\bfseries Qwen3-8B} \\[1pt]
\quad Zero-shot\frozen & 0.098 & 0.097 & 0.059 & 0.086 & \textbf{$-$79\%} \\
\quad Base harness\trained & 0.431 & 0.419 & 0.352 & 0.403 & --- \\
\quad DeepAgents\trained & 0.465 & 0.456 & 0.361 & 0.430 & \textbf{$+$7\%} \\
\quad INSPO\trained & 0.477 & 0.441 & 0.347 & 0.425 & \textbf{$+$5\%} \\
\quad EvolveR\trained & 0.443 & 0.450 & 0.353 & 0.418 & \textbf{$+$4\%} \\
\quad HarnessForge\trained & 0.430 & 0.395 & 0.274 & 0.370 & \textbf{$-$8\%} \\
\quad SkillRL\trained & 0.478 & 0.428 & 0.362 & 0.426 & \textbf{$+$6\%} \\
\rowcolor{oursrow}\quad + \Compression\trained & 0.453 & 0.461 & 0.329 & 0.418 & \textbf{$+$4\%} \\
\rowcolor{oursrow}\quad + \Steering\trained & 0.483 & \textbf{0.487} & 0.385 & 0.455 & \textbf{$+$13\%} \\
\rowcolor{oursrow}\quad \ours\trained & \textbf{0.492} & 0.473 & \textbf{0.409} & \textbf{0.460} & \textbf{$+$14\%} \\
\addlinespace[1pt]
\multicolumn{6}{@{}l}{\bfseries GPT-5-mini} \\[1pt]
\quad Zero-shot\frozen & 0.404 & 0.389 & 0.298 & 0.366 & --- \\
\bottomrule
\end{tabular*}
\caption{\textbf{CodeSearch validation, F1 averaged over the three
granularities.} $\Delta$ is the relative change in \emph{Overall} over the base
harness of the same model, and bold is best in a column within a model. Shaded rows are
ours. \trained\ marks an RL-trained model and \frozen\ one that was not fine-tuned.}
\label{tab:cs_main}
\end{table}
\fi
\label{sec:setup}

\paragraph{Environments.}
In \textbf{CodeSearch}~\citep{sutawika2026codescout} the agent localizes a bug in a real
repository by naming the file, module and entity to edit. We train on SWE-Smith and evaluate
on SWE-Bench Lite, Verified and Pro, scoring the mean F1 of the three names on $[0,1]$. In
\textbf{BIRD SQL}~\citep{li2023bird} the agent queries a real database to answer a
natural-language question. We train on the BIRD training split and evaluate on its dev split.
In \textbf{ALFWorld}~\citep{feng2025gigpo}, a text-world household environment, the agent
completes a household task through a sequence of actions. We train on the training games
and evaluate on the held-out seen and unseen games. In all three, evaluation uses tasks never
seen in training. $\Dhold$ is held out from the training split and is disjoint from every evaluation set.

\paragraph{Methods.}
Every method shares the environment, the policy and the training configuration, and
differs only in what surrounds the model. ``Base harness'' is the hand-written scaffold,
unchanged throughout training; ``$+$ \Compression'' and ``$+$ \Steering'' add one learned
program each and nothing else; \ours\ runs both, with \Compression\ learned once and
\Steering\ evolving with the policy.
The baselines take one representative of each component that can be learned alongside a
policy, namely memory, skills, prompts and the harness itself.
EvolveR~\citep{wu2025evolver} learns \emph{memory}, retrieving from an experience base
distilled from past rollouts. SkillRL~\citep{xia2026skillrl} learns \emph{skills},
inducing a reusable library of them. INSPO~\citep{zhou2025inspo} learns \emph{prompts},
co-evolving task instructions with the policy.
HarnessForge~\citep{chen2026harnessforge} learns the \emph{harness}, editing planning,
action and memory modules from observed faults.
DeepAgents~\citep{langchain2026deepagents} is the fixed-scaffold control, an elaborate
hand-written harness with a planning checklist, a scratchpad and sub-agents.

\newcommand{\figloopcaption}{\caption[F1 against inference time, Qwen3-8B on CodeSearch.]{\textbf{F1 against
inference time, Qwen3-8B on CodeSearch.} Each panel is one context budget. F1 is averaged over the
three granularities and over all validation tasks, as in \cref{tab:cs_main}, and inference time
is the average seconds an episode takes end to end.}}
\newcommand{\figrqcaption}{\caption{\textbf{\Compression\ against the base harness.}
Qwen2.5-7B on ALFWorld and Qwen3-8B on CodeSearch. Prefill tokens are the prompt tokens not served from the prefix cache, summed over an episode and measured by replaying the recorded episodes (\cref{app:kvcache}).}}

\ifonecolumn
\begin{figure}[t]
\centering
\begin{minipage}[t]{0.485\textwidth}\centering
\includegraphics[width=\linewidth]{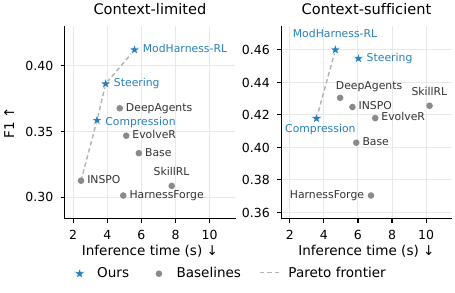}
\figloopcaption
\label{fig:loop}
\end{minipage}\hfill
\begin{minipage}[t]{0.485\textwidth}\centering
\includegraphics[width=\linewidth]{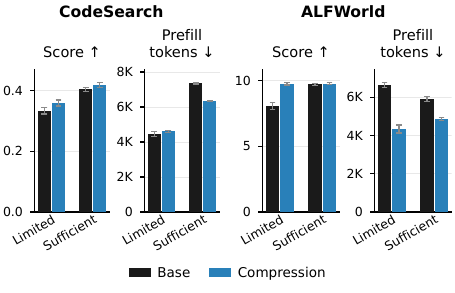}
\figrqcaption
\label{fig:rq1}
\end{minipage}
\end{figure}
\else
\begin{figure}[t]
\centering
\includegraphics[width=\arxivcolw]{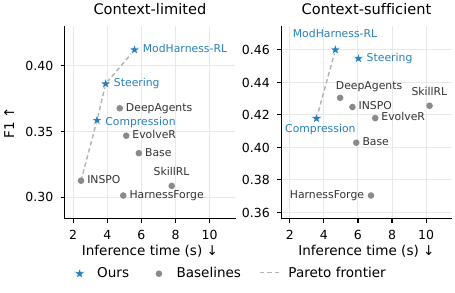}%
\figloopcaption
\label{fig:loop}
\end{figure}
\fi

\paragraph{Models and training.}
CodeSearch uses Qwen3-4B and Qwen3-8B and BIRD SQL uses Qwen3.5-2B and Qwen3.5-4B, both
with reasoning disabled, and ALFWorld uses Qwen2.5-1.5B-Instruct and Qwen2.5-7B-Instruct.
Qwen2.5-1.5B is warm-started with supervised fine-tuning before RL (\cref{app:details}).
We choose the model family for each environment so that RL training shows a clear signal, as with other choices the model either barely learns with RL or saturates.
Every method trains with GRPO under an identical configuration for 200 steps on one node of
eight H100s, and every metric is computed using the final checkpoint. The $K=20$ \Steering\
rounds are spaced evenly over training and run beside it, so the trainer never waits for
them. For a fair comparison, every method that learns from rollouts uses GPT-5-mini as its
creator. It writes the programs of \ours, the memory of EvolveR, the skills of SkillRL, the
prompts of INSPO and the harness edits of HarnessForge.

\paragraph{Context regimes.}
To evaluate the effect of the harness under different context budgets, we set a \emph{history budget}, the token allowance for accumulated
interaction, and run every environment and model under two budgets. We call one model
under one budget a \emph{setting}. In a \textbf{context-limited} setting, the interaction
history is capped at a token limit that runs out partway through a typical episode in every
environment, so the harness drops or compacts earlier turns. In a \textbf{context-sufficient} setting, the token limit is almost never reached. The specific token limit depends on each setting, so we defer it to \Cref{app:budgets}.

\subsection{Main Results}
\label{sec:main_results}

\cref{tab:cs_main} reports the full CodeSearch results. \ours\ improves over the base
harness by \fCsOursAvg\% on average, and the gain holds at the file, module and entity level
(\Cref{tab:cs_gran}). It also beats zero-shot GPT-5-mini, the creator of its harness, in
every setting, so the gain is not distillation from the creator.
\Cref{fig:benchmarks} applies each program alone to the base harness, and we can see that the gain from the two programs differs by domain. On CodeSearch, \Steering\ improves over the base harness by
\fBenchCsSteer\% on average over the four settings, which is better than \Compression\ at \fBenchCsComp\%. On ALFWorld under a limited
budget, averaged over the two models, \Compression\ improves by \fBenchAlfLimComp\%, which is better than \Steering\ at \fBenchAlfLimSteer\%.
On BIRD SQL, the two are close with Qwen3.5-2B, and with Qwen3.5-4B, \Compression\ performs better.

In \cref{fig:loop}, we also report the inference time, the average seconds an episode takes
end to end, for Qwen3-8B, and show that every method on the frontier is ours except INSPO under
the limited budget, which is also fast but scores below the base harness. \Cref{sec:rq1}
further examines these gains.

\ifonecolumn\else
\begin{figure}[t]
\centering
\includegraphics[width=\arxivcolw]{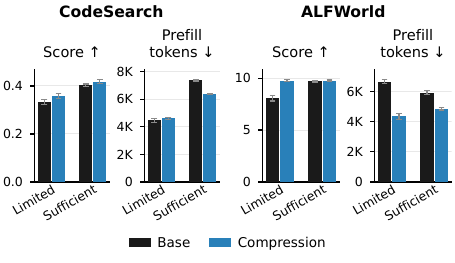}
\figrqcaption
\label{fig:rq1}
\end{figure}
\fi

\subsection{Analysis}
\label{sec:analysis}
\mbox{}\par\vspace{-\baselineskip}
\ifonecolumn\begin{wraptable}[16]{r}{0.5\textwidth}
\vspace{-10pt}
\wraptab{\begin{table}[t]
\centering
\scriptsize
\setlength{\tabcolsep}{2.6pt}
\begin{tabular*}{\arxivcolw}{@{\extracolsep{\fill}}lcccc@{}}
\toprule
Method & Score\,$\uparrow$ & Calls\,$\downarrow$ & Tokens\,$\downarrow$ & ms/turn\,$\downarrow$ \\
\midrule
\multicolumn{5}{@{}l}{\itshape CodeSearch $\cdot$ Qwen3-4B $\cdot$ Context-limited} \\
\quad GPT Compaction & 0.340 & 12,365 & 62.6M\textsuperscript{e} & 749\textsuperscript{e} \\
\quad \Compression & \textbf{0.345} & \textbf{36\textsuperscript{s}} & \textbf{0.183M\textsuperscript{e,s}} & \textbf{0.30} \\
\addlinespace[2pt]
\multicolumn{5}{@{}l}{\itshape CodeSearch $\cdot$ Qwen3-8B $\cdot$ Context-limited} \\
\quad GPT Compaction & \textbf{0.377} & 9,607 & 32.2M\textsuperscript{e} & 329\textsuperscript{e} \\
\quad \Compression & 0.359 & \textbf{36\textsuperscript{s}} & \textbf{0.183M\textsuperscript{e,s}} & \textbf{0.24} \\
\addlinespace[2pt]
\multicolumn{5}{@{}l}{\itshape ALFWorld $\cdot$ Qwen2.5-1.5B\textsuperscript{\dag} $\cdot$ Context-limited} \\
\quad GPT Compaction & 8.723 & 247,347 & 1,789.5M\textsuperscript{e} & 4,127\textsuperscript{e} \\
\quad \Compression & \textbf{8.893} & \textbf{38\textsuperscript{s}} & \textbf{0.198M\textsuperscript{e,s}} & \textbf{1.42} \\
\addlinespace[2pt]
\multicolumn{5}{@{}l}{\itshape ALFWorld $\cdot$ Qwen2.5-7B $\cdot$ Context-sufficient} \\
\quad GPT Compaction & \textbf{9.866} & 7,123 & 124.8M\textsuperscript{e} & $\ge$103\textsuperscript{e,b} \\
\quad \Compression & 9.757 & \textbf{38\textsuperscript{s}} & \textbf{0.198M\textsuperscript{e,s}} & \textbf{0.51} \\
\bottomrule
\end{tabular*}
\caption{\textbf{What each mechanism spends over one training run.} 
\emph{Score} is on the held-out validation set and \emph{ms/turn} is wall clock added per agent turn. Superscripts mark figures that were estimated (\textsuperscript{e}), scaled
(\textsuperscript{s}) or a lower bound (\textsuperscript{b}), as detailed in
\cref{app:cost}. \textsuperscript{\dag}SFT warm start.}
\label{tab:cost}
\end{table}
}
\vspace{-8pt}
\end{wraptable}\fi
\pp{The gain from \Compression}
\label{sec:rq1}
To investigate the gain from \Compression, we consider how it improves the base harness in
both performance and efficiency. We measure performance with the score of each benchmark and
efficiency with prefill tokens. Counting the prompt tokens alone would favour any method that
rewrites the history, because a rewritten history can break prefix-cache reuse, and a cached
prefix costs the server nothing to recompute. We therefore count prefill tokens, the prompt
tokens an episode sends multiplied by the share the prefix cache misses, summed over its model
calls. We measure them by replaying the recorded episodes with no turn dropped, so the base
harness keeps its full cache reuse but is not held to the budget (\cref{app:kvcache}).

\Cref{fig:rq1} shows that \Compression\ improves the base harness in two ways, depending on
the context regime. Under a limited budget, performance tends to improve, with \fCsLimGain\%
improvement for Qwen3-8B on CodeSearch and \fAlfLimGain\% improvement for Qwen2.5-7B on
ALFWorld. Under a sufficient budget, \Compression\ consistently reduces the prefill tokens.
Specifically, under the budget used in training, the prompt at an episode's last model call is
within \fLimCtxWithin\% between the two arms under a limited budget, and \Compression\ uses
\fAlfSufSave\% fewer tokens on ALFWorld and \fCsSufSave\% fewer on CodeSearch under a sufficient
one. Even though rewriting the history can lose some prefix-cache reuse, \Compression\ still
uses fewer prefill tokens: because \compress\ is a deterministic program, earlier turns keep an
identical rendering, so the majority of its cache stays valid.

\ifonecolumn\else\fi

\pp{Comparison with compaction by a strong teacher model}
\label{sec:rq2}
A natural alternative to compiling a program is standard compaction, which calls a teacher
model to rewrite the history whenever the budget fills. We use GPT-5-mini, the same model
that writes \Compression. This is a strong baseline, because the teacher is utilized online in every episode and adapts to the content of the episode. In comparison, \Compression\ uses the teacher once, offline, to
write a deterministic program that every episode then runs without a model call. \Compression{} scores
higher than compaction in \fCmpAhead\ of the \fCmpCells\ settings of \cref{tab:cost} and up to
\fCmpBehind\% lower in the other \fCmpBehindN, with on the order of \fCmpCalls$\times$
reduction in teacher-model calls and tokens over a training run in our estimates (\cref{app:cost}) and, in our estimates, greater than \fCmpLat$\times$ reduction in wall clock time per turn in the context-limited settings, showcasing the efficiency of the compression program.

\ifonecolumn\else\begin{table}[t]
\centering
\scriptsize
\setlength{\tabcolsep}{2.6pt}
\begin{tabular*}{\arxivcolw}{@{\extracolsep{\fill}}l*{4}{c}@{}}
\toprule
 & \multicolumn{2}{c}{Context-limited} & \multicolumn{2}{c}{Context-sufficient} \\
\cmidrule(lr){2-3} \cmidrule(lr){4-5}
Pipeline & 4B & 8B & 4B & 8B \\
\midrule
w/o generation & 0.302 & 0.378 & 0.365 & 0.375 \\
w/o judging & 0.356 & 0.377 & 0.376 & 0.411 \\
policy first, then harness & 0.344 & 0.395 & 0.290 & 0.444 \\
harness first, then policy & 0.341 & 0.381 & 0.352 & 0.451 \\
evolving \Compression & 0.336 & 0.279 & 0.360 & 0.398 \\
\rowcolor{oursrow}\ours & \textbf{0.368} & \textbf{0.412} & \textbf{0.424} & \textbf{0.460} \\
\bottomrule
\end{tabular*}
\caption[Ablation of \ours.]{\textbf{Ablation of \ours, CodeSearch validation F1 (Overall), Qwen3-4B and Qwen3-8B.} Same
metric as \cref{tab:cs_main}. Bold is best in a column, and the shaded row is the full system.}
\label{tab:ablation}
\end{table}
\fi

\pp{Ablation of \ours}
\label{sec:rq3}
\Cref{tab:ablation} ablates each component of \ours. Each component improves performance.

Without generation, the creator proposes no new triggers during training, and the policy
trains with one \Steering\ program fixed in advance. Without judging, every proposed trigger
goes to the gate instead of only those the judge admits. Each lowers the score in every setting, so
neither the creator nor the judge carries the result alone.

\ifonecolumn\begin{wraptable}[10]{r}{0.5\textwidth}
\vspace{-15pt}
\wraptab{\begin{table}[t]
\centering
\scriptsize
\setlength{\tabcolsep}{2.6pt}
\begin{tabular*}{\arxivcolw}{@{\extracolsep{\fill}}l*{4}{c}@{}}
\toprule
 & \multicolumn{2}{c}{Context-limited} & \multicolumn{2}{c}{Context-sufficient} \\
\cmidrule(lr){2-3} \cmidrule(lr){4-5}
Pipeline & 4B & 8B & 4B & 8B \\
\midrule
w/o generation & 0.302 & 0.378 & 0.365 & 0.375 \\
w/o judging & 0.356 & 0.377 & 0.376 & 0.411 \\
policy first, then harness & 0.344 & 0.395 & 0.290 & 0.444 \\
harness first, then policy & 0.341 & 0.381 & 0.352 & 0.451 \\
evolving \Compression & 0.336 & 0.279 & 0.360 & 0.398 \\
\rowcolor{oursrow}\ours & \textbf{0.368} & \textbf{0.412} & \textbf{0.424} & \textbf{0.460} \\
\bottomrule
\end{tabular*}
\caption[Ablation of \ours.]{\textbf{Ablation of \ours, CodeSearch validation F1 (Overall), Qwen3-4B and Qwen3-8B.} Same
metric as \cref{tab:cs_main}. Bold is best in a column, and the shaded row is the full system.}
\label{tab:ablation}
\end{table}
}
\vspace{-8pt}
\end{wraptable}\fi
We then ablate co-evolution by running the two stages one after the other, interchanging the order across two experiments: policy optimization first under a fixed harness, followed by harness optimization under the final policy, and vice versa. Both orders reduce performance in every setting, demonstrating the value of co-evolution.

Finally, \ours\ fits \compress\ once against the base model (\cref{sec:coevolution}).
Evolving \Compression\ instead refits \compress\ every ten steps as the policy trains. Results show a dip in performance compared to \ours, suggesting the importance of keeping context stable during policy training.

\ifonecolumn\else
\begin{figure}[t]
\centering
\includegraphics[width=\arxivcolw]{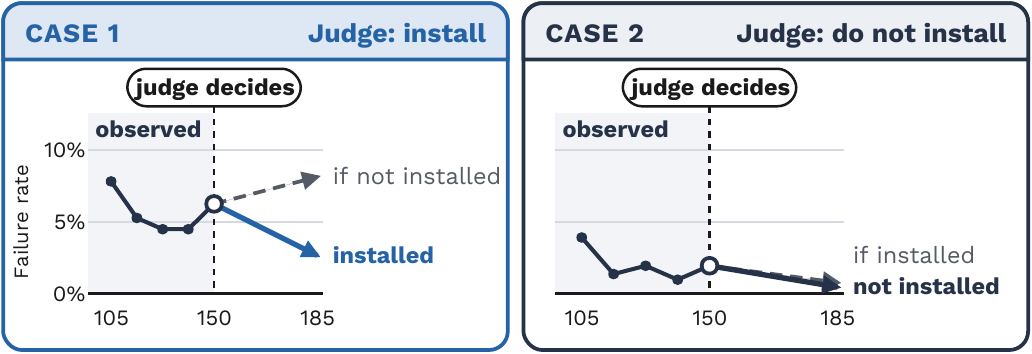}
\caption[The judge's decision and the failure rate after it.]{\textbf{The \Steering\ judge's decision in different cases on CodeSearch.} Dashed lines show the run the judge did not choose.}
\label{fig:gating_main}
\end{figure}
\fi

\vspace{1.2mm}\noindent\textbf{Does the judge know when \Steering\ is needed?}
Partway through CodeSearch training, the judge reads how often one failure has occurred so far and decides whether to install guidance for it. We continue training from that point both with and without the guidance (\cref{fig:gating_main}). In Case~1, the failure keeps recurring, and the judge installs the guidance, which cuts the failure. In Case~2, the failure is already rare and still falling, so the judge holds off, and indeed, without the guidance, the failure keeps falling as training continues. \Cref{fig:gating} gives the full version.

\ifonecolumn
\begin{figure}[t]
\centering
\begin{minipage}[b]{0.49\textwidth}\centering
\includegraphics[width=\linewidth]{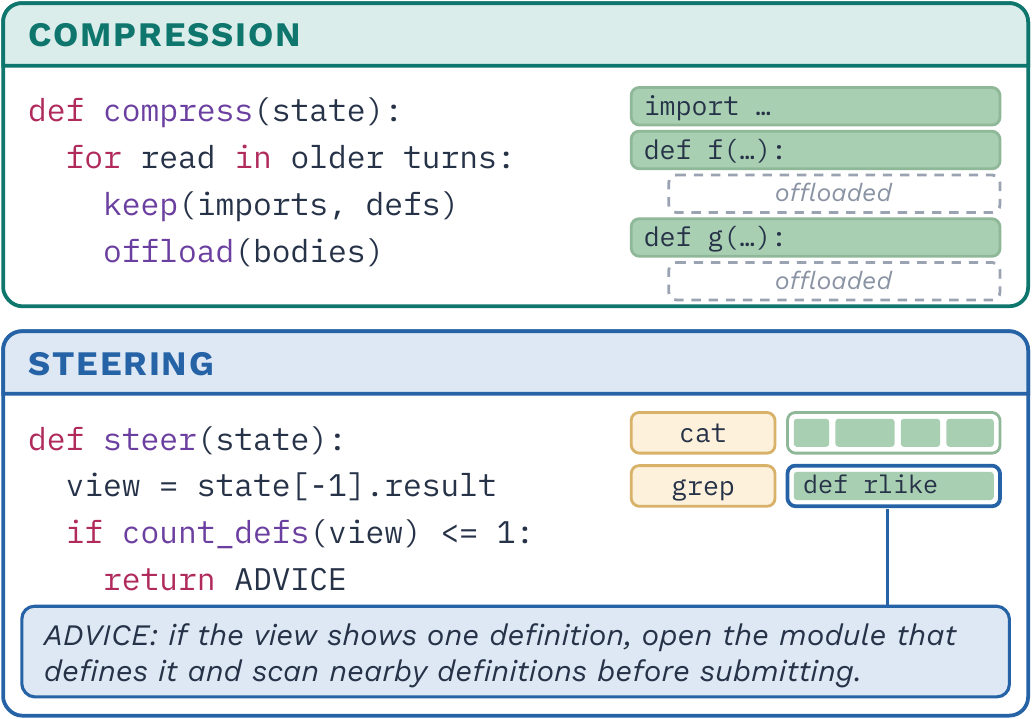}
\subcaption{}\label{fig:casestudy}
\end{minipage}\hfill
\begin{minipage}[b]{0.49\textwidth}\centering
\includegraphics[width=\linewidth]{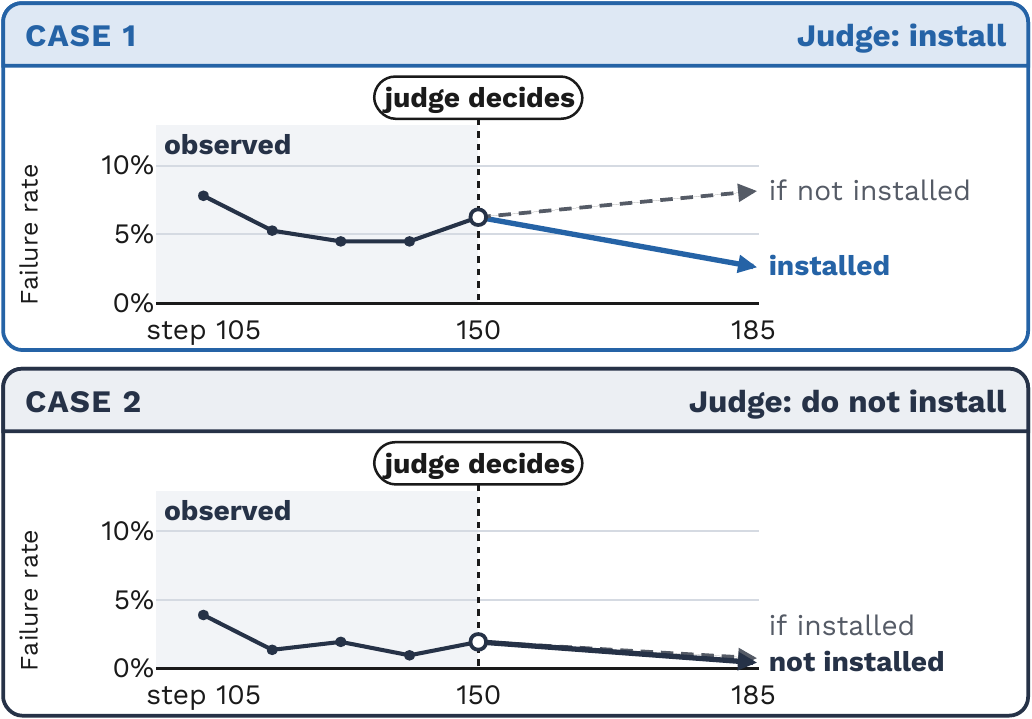}
\subcaption{}\label{fig:gating_main}
\end{minipage}
\caption{\textbf{(a)} \textbf{A learned \Compression\ and \Steering\ program} on CodeSearch, as high-level pseudocode. \textbf{(b)} \textbf{The \Steering\ judge's decision in different cases on CodeSearch.} Dashed lines show the run the judge did not choose.}
\end{figure}
\else
\begin{figure}[t]
\centering
\includegraphics[width=\arxivcolw]{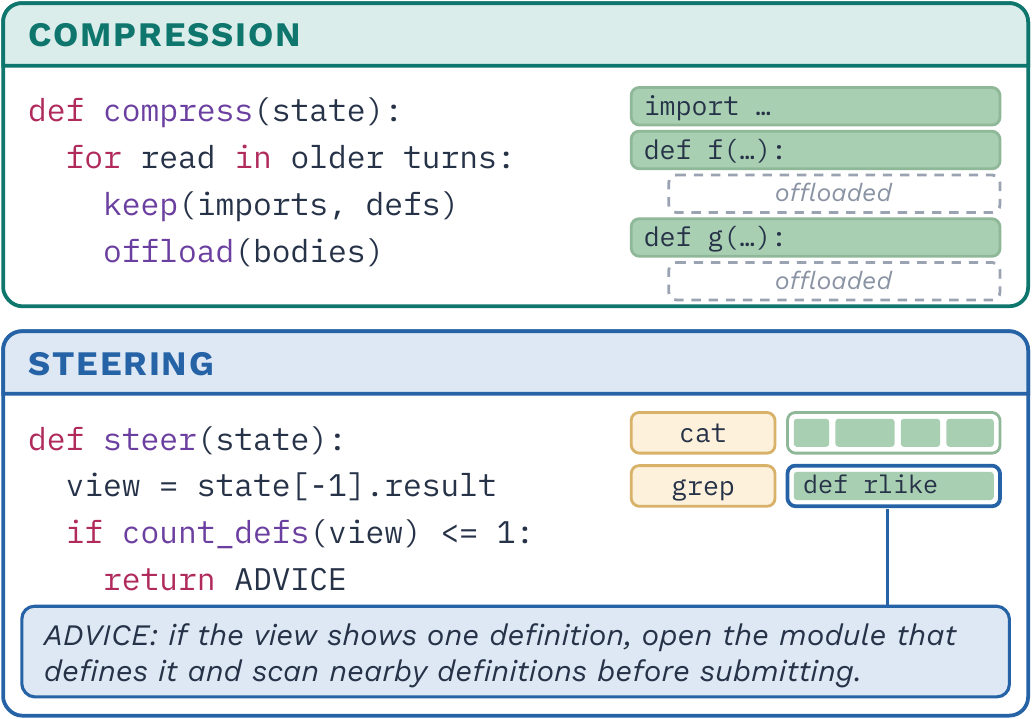}
\caption[A learned \Compression\ and \Steering\ program.]{\textbf{A learned \Compression\ and \Steering\ program} on CodeSearch, as high-level pseudocode.}
\label{fig:casestudy}
\end{figure}
\fi

\pp{Case studies}
\Cref{fig:casestudy} shows one learned \compress\ program and one learned \steer\ trigger as
high-level pseudocode. \Cref{app:programs} (\cref{fig:compress_example,fig:steer_example}) gives the details and applies both to 
episodes.

\section{Related Work}
\label{sec:related}

We discuss the two lines of work closest to ours
and defer the broader literature to \Cref{app:extended_related}.

\sparagraph{Harness and Context Optimization during Training}
A line of recent work updates the harness together with the weights, either by letting an
LLM agent rewrite the harness
code~\citep{chen2026coharness,chen2026harnessforge,hebbar2026sia,kim2026whale} or by
training the policy with RL to manage its own context~\citep{shao2026clm}.
These approaches rely on a model to rewrite the harness or to operate on the context
directly. \ours\
instead writes harness components as programs over typed blocks that capture rich
interaction, which run without a model call, keep the KV cache largely reusable and can be rerun
over stored trajectories.

\pp{Co-Evolving Memory, Skills and Prompts during Training}
Another line of work evolves what the agent reads together with the policy. It distills
experience into a memory~\citep{wu2025evolver,xie2026edge}, induces, writes or retires
skills and guidance~\citep{xia2026skillrl,ge2026skillforge,wang2025sage,qi2026vace}, or
evolves task instructions~\citep{zhou2025inspo}, with some methods withdrawing this guidance as
the policy improves~\citep{shi2026pats,xie2026edge}. These methods only append to the
prompt, and their guidance is retrieved by semantic similarity, matched to the task type
or selected by a model.
\ours\ instead conditions its guidance on typed blocks, so each trigger can be replayed and
targeted at a specific failure, enabling more targeted and deterministic control.

\section{Conclusion}
\label{sec:conclusion}

\ours\ treats the harness of agentic RL as programs that are learned from data and evolved
with the model. It records the interaction as \StateN{} and writes two
harness components as programs over these blocks, \Compression, which builds the model's
context, and \Steering, which supplies guidance. During training, \ours\ proposes candidate
programs from recent trajectories, verifies them by execution, and replays them over past
trajectories. Across three environments and two context budgets, each learned program
improves over the fixed-harness baseline in nearly every setting, raising the score when context is scarce and
reducing prefill tokens when it is not. On CodeSearch it also outperforms memory, skill,
prompt and harness evolution baselines. 

\label{probe:endofcontent}

\section*{Limitations}

\paragraph{Scope of the comparison.}
We investigate CodeSearch most thoroughly, running the full method and the third-party
baselines under both a limited and a sufficient budget with two model sizes, to establish the
signal. ALFWorld and BIRD SQL then test whether each program generalizes to other
environments, again under both budgets and with two model sizes. Ideally the full method and
baselines would also cover ALFWorld and BIRD SQL.

\paragraph{Gate sample size.}
The gate scores each candidate on a limited set of held-out training tasks, so it can
occasionally accept a program that is slightly worse. A larger held-out set would make the
gate more reliable, at a higher cost.

\paragraph{Proprietary creator.}
The programs are written by a proprietary model, GPT-5-mini. To enable a fair comparison, every
learning baseline uses the same creator. GPT-5-mini writes the \Compression\ and \Steering\
programs of \ours, the memories, skills, prompts and harness edits of the baselines, and the
summaries of the compaction baseline. For \ours, the policy sees only the output of executing
these programs (\cref{app:programs} shows examples). This text is
never a loss target, and the programs of \ours\ are built once and can then be reused.

\paragraph{What deterministic programs can capture.}
The harness is a set of deterministic programs over $s_t$, so it captures only patterns that
an explicit rule over the recorded interaction can detect and act on, such as a repeated tool
call or a tool result whose structure shows what to keep. It suits interaction that is
structured and verifiable, such as tool calls with typed results, and is less effective for
open-ended interaction such as conversation, where the relevant signal is semantic and hard to
detect with a rule.

\section*{Ethical considerations}

This work trains and evaluates language model agents on publicly released benchmarks for code
localization, embodied household tasks and text-to-SQL. No other personal data was collected or
used, and it involves no human subjects. The harness programs are generated by a
language model and executed in the training loop, so they are compiled and run under
a restricted policy before installation rather than executed as arbitrary code.

\bibliography{refs}
\bibliographystyle{plainnat}

\appendix
\crefalias{section}{appendix}
\crefalias{subsection}{appendix}
\crefalias{subsubsection}{appendix}
\section{Experimental Details}
\label{app:details}

\subsection{Environments and Training}
\label{app:training}

\pp{Shared setup}
All runs train with GRPO~\citep{shao2024deepseekmath} in verl~\citep{sheng2025hybridflow},
and vLLM~\citep{kwon2023vllm} serves the policy for training and evaluation. Each run uses
FSDP with parameter and optimizer offloading for the policy, and a tensor parallel size of
one for the policy server. Within an environment, every method
uses the same model, training data apart from $\Dhold$, agent loop, tools, reward and
optimization settings as the base harness, so the methods differ only in the harness around the model. We take the
optimization settings of each environment from the work we follow for that environment, as
given below, and keep them unchanged for every method and both context regimes. Every run
trains for 200 steps and is evaluated at its step-200 checkpoint. Every method that writes
programs, memories, skills, instructions or harness edits uses GPT-5-mini to write them, accessed through the Azure OpenAI Service. We note that the models train slowly with reasoning
enabled, so in every environment we suppress reasoning and let the model write a short
rationale before each action.

\pp{Training objective}
\label{app:grpo}
For each task $q$, GRPO~\citep{shao2024deepseekmath} samples a group of $G$
trajectories $\{\tau_i\}_{i=1}^{G}$ from the current policy and normalizes their
rewards $R_i = r(\tau_i)$ within the group to form advantages,
\begin{equation}
\hat{A}_i = \frac{R_i - \operatorname{mean}(\{R_j\}_{j=1}^{G})}
                 {\operatorname{std}(\{R_j\}_{j=1}^{G})}.
\label{eq:grpo_adv}
\end{equation}
The policy is updated by the clipped surrogate
\begin{equation}
\begin{split}
\mathcal{L}_{\text{GRPO}}(\theta) =
{}&\frac{1}{G}\sum_{i=1}^{G}\frac{1}{|\tau_i|}\sum_{t=1}^{|\tau_i|}
\min\!\Big(\rho_{i,t}\,\hat{A}_i,\\
&\qquad \operatorname{clip}(\rho_{i,t}, 1-\epsilon, 1+\epsilon)\,\hat{A}_i\Big)\\
&{}-\; \beta\, \mathbb{D}_{\text{KL}}\!\left[\pi_\theta \,\|\, \pi_{\text{ref}}\right],
\end{split}
\label{eq:grpo_loss}
\end{equation}
where $\rho_{i,t}$ is the token-level importance ratio, $\epsilon$ the clip range,
$\beta$ the KL weight, and $\pi_{\text{ref}}$ a fixed reference policy.

\pp{CodeSearch}
We follow the setup of CodeScout~\citep{sutawika2026codescout}. Each task gives the agent a
repository checked out before a fix, together with the issue that the fix resolves, and
asks it to name the files, modules and entities that the fix changes. We train on tasks
from SWE-Smith~\citep{yang2025swesmith} and evaluate on 288 tasks, 100 from SWE-Bench Lite,
100 from SWE-Bench Verified~\citep{jimenez2024swebench} and 88 from SWE-Bench
Pro~\citep{deng2025sweprod}.
We use Qwen3-4B and Qwen3-8B. Each step samples 8 tasks and 8 episodes per task, and makes
one policy update on the whole batch. We use a constant learning rate of $10^{-6}$ with no
warmup, a low-variance KL loss with coefficient $0.001$, no entropy bonus, and a sampling
temperature of $1.0$ during training. The policy generates at most 1{,}024 tokens per turn
and takes at most 10 turns per episode, with one tool call per turn. A tool call is a shell
command that times out after 30 seconds, and the agent observes at most 10{,}000 bytes of
its output. The reward is the sum of the F1 scores of the predicted files, modules and
entities against those the gold patch changes, which lies in $[0,3]$. We report the mean of
the three F1 scores, which lies in $[0,1]$.

\pp{ALFWorld}
We follow the setup of GiGPO~\citep{feng2025gigpo}. Each game places the agent in a
text-rendered household with a goal from one of six task types~\citep{shridhar2021alfworld}. After
each action the agent observes the result of the action and the list of actions that are
admissible next. We train on the training games and evaluate on the 274 evaluation games,
140 from the seen split and 134 from the unseen split.
We use Qwen2.5-1.5B-Instruct and Qwen2.5-7B-Instruct.
Before RL, Qwen2.5-1.5B-Instruct wins none of the ALFWorld training games, so every GRPO
group has zero advantage and gives no gradient. We therefore warm-start it with supervised
fine-tuning on 288 games won by ALFWorld's built-in rule-based expert (9 steps of 32 games), and
RL then proceeds identically for every arm.
Each step samples 16 games and 8
episodes per game, and makes one policy update on the whole batch. We use a learning rate
of $10^{-6}$, a low-variance KL loss with coefficient $0.01$, an entropy coefficient of
$0.001$, and a sampling temperature of $0.6$ during training. The policy generates at most
2{,}048 tokens per turn and takes at most 50 turns per episode. The reward is $1$ when the
game is won and $0$ otherwise, with no penalty for invalid actions. Unlike GiGPO, which
shows the policy only the two most recent turns, we keep the full history of the episode in
the prompt and let it grow until it reaches the context limit, so that training matches a
multi-turn setting in which the history accumulates.

\pp{BIRD SQL}
We follow the setup of MTIR-SQL~\citep{xu2025mtirsql}. Each task gives the agent a
natural-language question, the full schema of its database and the evidence that BIRD
provides with the question, which we place before the question following
SQL-R1~\citep{ma2025sqlr1}. We train on the BIRD training split and evaluate on the
1{,}534 questions of the dev split. We drop the 663 training questions whose task statement
alone exceeds 6{,}144 tokens, which leaves 7{,}726 questions for training. We use
Qwen3.5-2B and Qwen3.5-4B. Each step samples 64 questions and 5 episodes per question, and
makes one policy update on the whole batch. We use a constant learning rate of $10^{-6}$,
no KL loss, no entropy bonus
and a sampling temperature of $0.6$ during training. The policy takes at most 6 turns per
episode and issues one SQL query per turn. The agent observes the column headers of each
query result and at most 10 of its rows. The reward is $1$ when the final query returns the
same result as the gold query and $0$ otherwise.

\subsection{Baselines}
\label{app:baselines}

We implement every baseline on our trainer, so each uses the model, data, agent loop,
tools, history budget and optimization settings of \ours, with GPT-5-mini wherever the
method calls an LLM. All baselines are run on CodeSearch.

\pp{EvolveR}
Following \citet{wu2025evolver}, GPT-5-mini distills principles from finished episodes
into an experience base during training, and the principles retrieved for a task are added
to its context.

\pp{SkillRL}
Following \citet{xia2026skillrl}, each skill has a title, a principle and a condition for
when to apply it. General skills are always shown, together with the six task-specific
skills that best match the task. Every five validation steps, GPT-5-mini adds at most three
new skills for each task category that scores below 40\%.

\pp{INSPO}
Following \citet{zhou2025inspo}, we port the authors' implementation. A population of
instructions is placed after the task statement. During training, an instruction is drawn
for each episode in proportion to its importance weight, and at evaluation the instruction
with the highest weight is used. Weak instructions are removed by successive halving, and
GPT-5-mini writes new instructions by reflecting on failed episodes.

\pp{HarnessForge}
HarnessForge~\citep{chen2026harnessforge} generates an executable harness made of
planning, action and memory modules, and then aligns the policy to the selected harness by
supervised fine-tuning on trajectories collected under it. Its released code cannot be run
on CodeSearch as is. Each benchmark it reports is connected by hand, with its own tool
runtime, its own tool loader that requires a sample of that benchmark, and its own scorer,
and a generated harness is a Python package built on the agent and provider classes of
that code, which call tools through a function-calling interface. We therefore port the
loop that evolves the harness and leave out the benchmark-specific parts. Every 50
training steps, which gives three rounds, GPT-5-mini reads training episodes, attributes the
most frequent fault to the planning, action or memory module, writes an improvement brief
for that module, and proposes an edit, following the first two stages of HarnessForge. We
serve a planning or action edit as guidance at a turn, and a memory edit as a rewrite of
the interaction history. The policy trains with the same GRPO as every other method in
place of the fine-tuning stage.

\pp{DeepAgents}
We use DeepAgents~\citep{langchain2026deepagents} version 0.6.12 with its default
components, a planning checklist, a virtual file system the agent can write notes to,
sub-agents that run a task in a fresh context, and summarization that moves old context to
files when the context grows. It is fixed during training.

\subsection{\ours\ Settings}
\label{app:settings}

\ours\ fits \compress\ once before training and runs $K=20$ \Steering\ rounds, one every
10 training steps, alongside the trainer (see \cref{app:cadence} for a sensitivity analysis). GPT-5-mini at medium reasoning effort writes
every program and serves as every judge. Both parts score a candidate against the
incumbent with the same policy, on the same $\Dhold$ tasks and with the same agent loop,
tools and history budget as training, so a difference in score reflects the program and not
a change in the policy or the sample.

\pp{Held-out training tasks}
Both parts of \ours\ choose among programs on $\Dhold$, a set of tasks held out from the
training split. We draw $\Dhold$ once per environment with a fixed seed and separate tasks
by identity, using the issue text in CodeSearch, the game file in ALFWorld, and the
database and question in BIRD SQL. The policy of \ours\ trains on the remaining tasks, so
no task is both in $\Dhold$ and trained on. $\Dhold$ has 128 tasks in CodeSearch and
ALFWorld. In BIRD SQL it has 128 questions in the context-sufficient settings and 256 in
the context-limited settings, where the 256 contain the 128. No program is chosen on an
evaluation set, and $\Dhold$ is disjoint from every evaluation set we report.

\subsection{History Budget and Overflow}
\label{app:overflow}

We follow the design of verl-agent~\citep{feng2025gigpo}, which drops earlier turns from
the history and tells the policy that they existed. The history budget applies to the interaction history after \compress\ has run. When the
compressed history still exceeds the budget, the harness drops the oldest whole turns until
it fits, always keeps the task statement, and marks the gap with the line
\texttt{[earlier turns omitted]}. Unlike verl-agent, which always keeps a fixed number of
recent turns, we drop turns only when the history exceeds the budget. The
rule applies to every method, the base harness included, so the budget is binding on all
of them in the same way. This is the dynamic eviction referred to in \cref{app:kvcache}.

\section{Extended Related Work}
\label{app:extended_related}

\pp{Multi-turn agent learning}
ReAct established the interleaving of language reasoning and environment actions
\citep{yao2023react}; AgentBench made the breadth of such interactions visible across
environments~\citep{liu2024agentbench}. Search-R1, WebAgent-R1, Agent-RLVR, RAGEN, and
GiGPO train multi-turn tool policies with verifiable outcomes or trajectory-aware
objectives~\citep{jin2025searchr1,wei2025webagentr1,da2025agentrlvr,
wang2025ragen,feng2025gigpo}. Toolformer, ToolLLM, and ToolRL study tool acquisition at
different scales~\citep{schick2023toolformer,qin2023toolllm,qian2025toolrl}. These
methods primarily improve policy parameters or objectives. \ours\ is complementary. It
represents the harness as programs over a structured \StateN, and learns and evolves them alongside agent learning.

\pp{Context as a managed resource}
Agent-computer interfaces can dominate software-agent performance~\citep{yang2024sweagent,
xia2024agentlessdemystifyingllmbasedsoftware}. Position sensitivity and context rot show that nominal context capacity is not the same
as usable evidence~\citep{liu2024lostmiddle,chroma2025contextrot}. MemGPT treats context
as a memory hierarchy, ReSum periodically summarizes search trajectories, Recursive
Language Models delegate work over long inputs, and Context-1 trains self-editing search
behavior~\citep{packer2024memgpt,yu2025resum,rlm2025,bashir2026context1}. These systems
motivate our \Compression{} abstraction. 
\ours\ writes \compress\ as a program over typed blocks, which runs without a
model call and keeps the KV cache largely reusable. We fit \compress\ once before training and
hold it fixed.

\pp{Scaffolds and rubrics}
The classical scaffolding view supplies support that can be withdrawn as competence
grows~\citep{wood1976role,vygotsky1978mind}. Rubric-scaffolded RL supplies externally specified reasoning criteria during exploration
and decays their use as training proceeds~\citep{qiu2025ruscarl}. 
\ours\ instead conditions its guidance on typed blocks, and its \Replay\ mechanism reruns
each candidate \Steering\ program over trajectories from past training steps, and a judge installs it
only for the failures that persist as the model trains.

\section{Additional Results}
\label{app:additional_results}

\begin{figure*}[t]
\centering
\includegraphics[width=\textwidth]{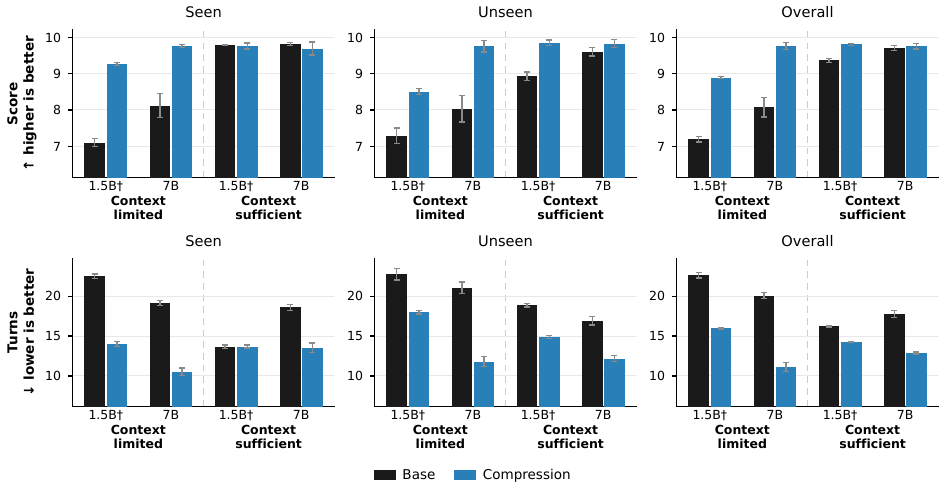}
\caption{\textbf{ALFWorld validation, base harness against \Compression.} Every model
is Qwen2.5 and the tick gives its scale. Whiskers are $\pm$1 SD over repeated evaluations of each trained checkpoint. \textsuperscript{\dag}SFT warm start (\cref{app:details}).}
\label{fig:alf_main}
\end{figure*}

\begin{figure*}[t]
\centering
\includegraphics[width=\textwidth]{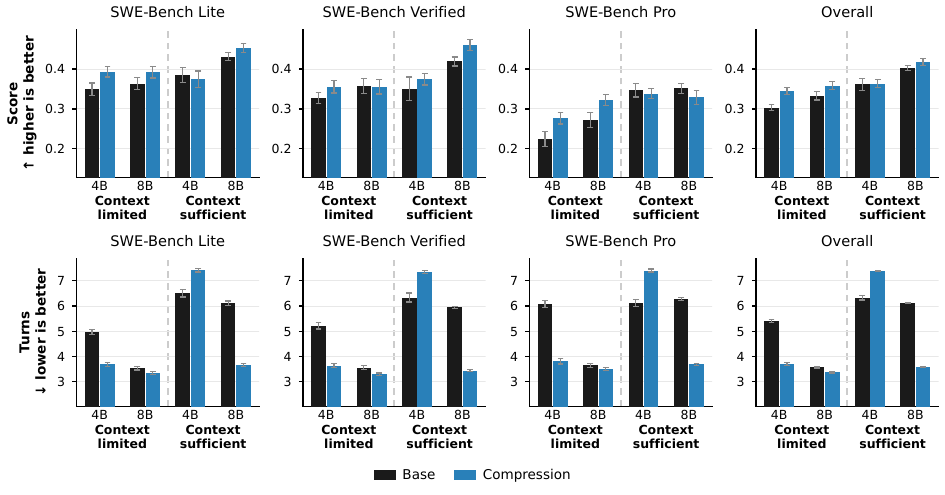}
\caption{\textbf{CodeSearch validation, base harness against \Compression.} Every model
is Qwen3 and the tick gives its scale, and \emph{Overall} is the instance weighted mean.
This is the per-split view of the aggregate in \cref{tab:cs_main}, where every number is
given exactly. Whiskers are $\pm$1 SD over repeated evaluations of each trained checkpoint.}
\label{fig:cs_main}
\end{figure*}

\begin{figure*}[t]
\centering
\includegraphics[width=\textwidth]{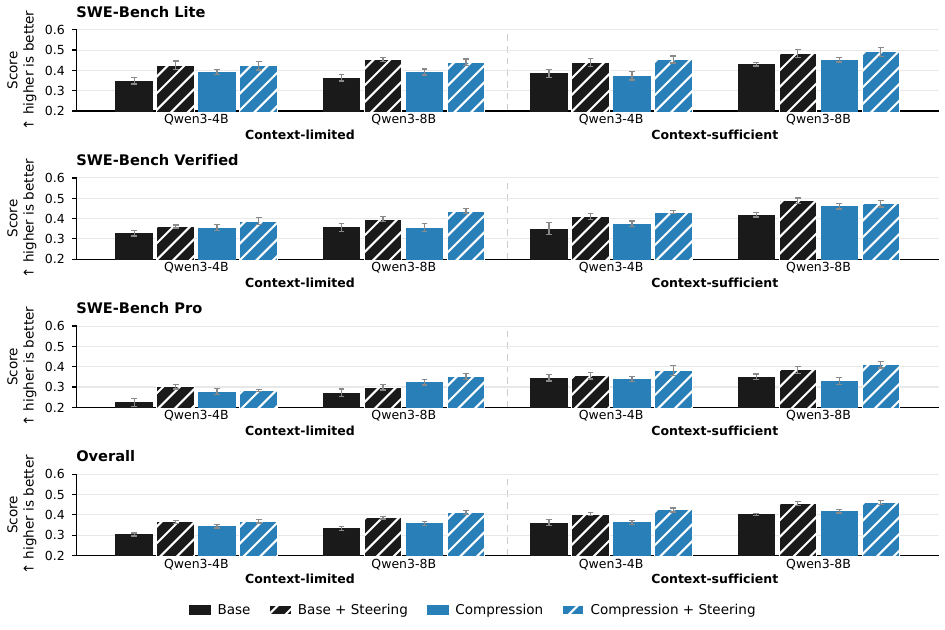}
\caption{\textbf{Learned \Steering\ on top of each context policy, CodeSearch overall
validation score.} \Compression\ $+$ \Steering\ is \ours. Whiskers are $\pm$1 SD over repeated evaluations of each trained checkpoint.}
\label{fig:ctrl_codesearch}
\end{figure*}

This appendix carries the context budget behind each regime, together with the
per-granularity (\cref{tab:cs_gran}), per-cell (\cref{fig:alf_main,fig:cs_main,fig:ctrl_codesearch}) and per-step views of the same runs reported in
\cref{sec:experiments}. Error bars
are $\pm$1 SD over ten evaluations of each trained checkpoint on CodeSearch and three on
ALFWorld and BIRD SQL.

\subsection{Sensitivity to the update interval}
\label{app:cadence}

\Steering\ runs $K=20$ rounds over 200 training steps, one every 10 steps. Like the other
hyperparameters, this interval is chosen in advance and not tuned. As a small check of its
sensitivity, we also train \ours\ on CodeSearch under a sufficient budget with an update every
5 steps ($K=40$) and every 20 steps ($K=10$). \Cref{tab:cadence} shows that an interval of 10
scores the highest with Qwen3-4B and ties with an interval of 20 with Qwen3-8B.

\begin{table}[h]
\centering
\small
\begin{tabular}{lccc}
\toprule
Update interval (steps) & 5 & 10 & 20 \\
\midrule
Qwen3-4B & .404 & \textbf{.424} & .404 \\
Qwen3-8B & .429 & \textbf{.460} & \textbf{.461} \\
\bottomrule
\end{tabular}
\caption{\textbf{Overall score of \ours\ on CodeSearch under a sufficient budget, by update interval.}}
\label{tab:cadence}
\end{table}

\subsection{Context budgets}
\label{app:budgets}

The \emph{history budget} is the token allowance for accumulated interaction, and it is
set per cell. It excludes the task description, so tasks with descriptions of different
lengths receive the same budget.
A \textbf{context-limited} cell sets it to 1,024 tokens, and a
\textbf{context-sufficient} cell to 8,192 tokens by default. On BIRD SQL, the 1,024-token budget runs
out on \fSqlFireLo\ to \fSqlFireHi\% of base-harness turns, dropping earlier turns, while
the 8,192-token budget never runs out. CodeSearch uses the same two budgets. ALFWorld
episodes run longer than 8,192 tokens, so its context-sufficient budget is raised to 12,288 tokens.

\begingroup
\subsection{Cost accounting}
\label{app:cost}

In \cref{tab:cost}, \emph{Score} is measured on the held-out validation set overall, tokens
are input plus output, \emph{ms/turn} is wall clock added to one episode divided by its number of turns, and bold is better
within a setting. \textsuperscript{e}~is estimated by differencing two runs,
\textsuperscript{s}~is scaled from a comparable campaign, and \textsuperscript{b}~is a lower
bound priced at the per-call cost of the setting where that cost is measurable.
\endgroup

\subsection{Prefix-cache reuse}
\label{app:kvcache}

\begin{figure*}[t]
\centering
\includegraphics[width=\textwidth]{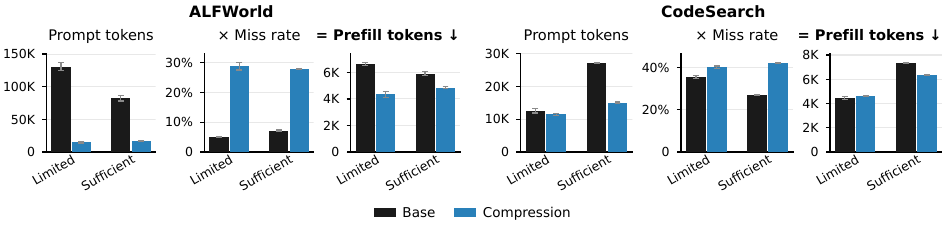}
\caption[Prefix-cache reuse under \Compression.]{\textbf{Prefix-cache reuse under
\Compression.}
Each benchmark multiplies the prompt tokens an episode sends, summed over its model calls,
by the share of them the cache misses.
Results are collected from Qwen2.5-7B on ALFWorld and Qwen3-8B on CodeSearch.
Whiskers are $\pm$1 SD over repeated evaluations of each trained checkpoint.}
\label{fig:kvcache}
\end{figure*}

A method that modifies the context can break prefix-cache reuse. An append-only harness
never changes a token it has already sent, so every prefix stays valid, whereas a rewritten
history appears to move under the cache. \Compression\ is a deterministic program, so an
earlier turn keeps an identical rendering unless the program changes it, and the cache stays
valid for every turn it leaves unchanged.
We measure reuse by replaying every validation episode with dynamic eviction removed from
both arms, which gives the base harness its best case. The prefill tokens in
\cref{fig:rq1} are the product drawn in \cref{fig:kvcache}, the prompt tokens an episode
sends times the share the cache misses. \Compression\ serves \fKvReuseLo\ to
\fKvReuseHi\% of prompt tokens from cache against \fKvBaseLo\ to \fKvBaseHi\% for the
append-only base, so the rate falls without collapsing, and because it is taken on far
fewer tokens the prefill tokens are lower in \fKvCheaper\ of \fKvCells\ cells, by up to
\fKvBestSave\%. Measured instead as the prompt at an episode's last model call, served
context is within \fLimCtxWithin\% between the arms under a limited budget, and
\Compression\ uses \fAlfSufSave\% fewer tokens on ALFWorld and \fCsSufSave\% fewer on
CodeSearch under a sufficient one.

\begin{table*}[t]
\centering
\scriptsize
\setlength{\tabcolsep}{2.2pt}
\ifdefined\gcwmin\else\newlength\gcwmin\newlength\gcwtmp\fi
\settowidth{\gcwmin}{\textbf{0.000}}
\settowidth{\gcwtmp}{SWE-Bench Lite ($n{=}100$)}\setlength{\gcwtmp}{\dimexpr\gcwtmp/3-2\tabcolsep\relax}\ifdim\gcwtmp<\gcwmin \setlength{\gcwtmp}{\gcwmin}\fi\edef\gcwa{\the\gcwtmp}
\settowidth{\gcwtmp}{SWE-Bench Verified ($n{=}100$)}\setlength{\gcwtmp}{\dimexpr\gcwtmp/3-2\tabcolsep\relax}\ifdim\gcwtmp<\gcwmin \setlength{\gcwtmp}{\gcwmin}\fi\edef\gcwb{\the\gcwtmp}
\settowidth{\gcwtmp}{SWE-Bench Pro ($n{=}88$)}\setlength{\gcwtmp}{\dimexpr\gcwtmp/3-2\tabcolsep\relax}\ifdim\gcwtmp<\gcwmin \setlength{\gcwtmp}{\gcwmin}\fi\edef\gcwc{\the\gcwtmp}
\settowidth{\gcwtmp}{Overall ($n{=}288$)}\setlength{\gcwtmp}{\dimexpr\gcwtmp/3-2\tabcolsep\relax}\ifdim\gcwtmp<\gcwmin \setlength{\gcwtmp}{\gcwmin}\fi\edef\gcwd{\the\gcwtmp}
\begin{tabular*}{\textwidth}{@{\extracolsep{\fill}}l*{3}{>{\centering\arraybackslash}p{\gcwa}}*{3}{>{\centering\arraybackslash}p{\gcwb}}*{3}{>{\centering\arraybackslash}p{\gcwc}}*{3}{>{\centering\arraybackslash}p{\gcwd}}cc@{}}
\toprule
 & \multicolumn{3}{c}{SWE-Bench Lite ($n{=}100$)} & \multicolumn{3}{c}{SWE-Bench Verified ($n{=}100$)} & \multicolumn{3}{c}{SWE-Bench Pro ($n{=}88$)} & \multicolumn{3}{c}{Overall ($n{=}288$)} &  &  \\
\cmidrule(lr){2-4} \cmidrule(lr){5-7} \cmidrule(lr){8-10} \cmidrule(lr){11-13}
Method & F & M & E & F & M & E & F & M & E & F & M & E & Overall & $\Delta$ \\
\midrule
\multicolumn{15}{@{}l}{\itshape Context-limited} \\[1pt]
\addlinespace[1pt]
\multicolumn{15}{@{}l}{\bfseries Qwen3-4B} \\[1pt]
\quad Zero-shot\frozen & 0.343 & 0.118 & 0.055 & 0.309 & 0.080 & 0.027 & 0.242 & 0.082 & 0.015 & 0.300 & 0.094 & 0.033 & 0.142 & \textbf{$-$53\%} \\
\quad Base harness\trained & 0.555 & 0.296 & 0.196 & 0.578 & 0.237 & 0.168 & 0.364 & 0.183 & 0.125 & 0.505 & 0.241 & 0.165 & 0.303 & --- \\
\quad DeepAgents\trained & 0.567 & 0.310 & 0.197 & 0.557 & 0.282 & 0.162 & \textbf{0.499} & 0.248 & 0.130 & 0.542 & 0.281 & 0.164 & 0.329 & \textbf{$+$9\%} \\
\quad INSPO\trained & 0.580 & 0.270 & 0.120 & 0.594 & 0.247 & 0.109 & 0.491 & 0.182 & 0.087 & 0.558 & 0.235 & 0.106 & 0.299 & \textbf{$-$1\%} \\
\quad EvolveR\trained & 0.530 & 0.280 & 0.172 & 0.554 & 0.274 & 0.154 & 0.436 & 0.159 & 0.063 & 0.510 & 0.241 & 0.132 & 0.294 & \textbf{$-$3\%} \\
\quad HarnessForge\trained & 0.578 & 0.332 & 0.206 & 0.587 & 0.296 & 0.170 & 0.484 & 0.252 & 0.130 & 0.553 & 0.295 & 0.170 & 0.339 & \textbf{$+$12\%} \\
\quad SkillRL\trained & 0.547 & 0.327 & 0.203 & 0.598 & 0.306 & 0.218 & 0.436 & \textbf{0.283} & \textbf{0.206} & 0.531 & 0.306 & 0.209 & 0.349 & \textbf{$+$15\%} \\
\rowcolor{oursrow}\quad + \Compression\trained & 0.584 & 0.352 & 0.244 & 0.566 & 0.293 & 0.208 & 0.428 & 0.242 & 0.161 & 0.530 & 0.298 & 0.206 & 0.345 & \textbf{$+$14\%} \\
\rowcolor{oursrow}\quad + \Steering\trained & \textbf{0.620} & 0.388 & 0.270 & 0.579 & 0.303 & 0.201 & 0.466 & 0.272 & 0.168 & 0.559 & \textbf{0.323} & 0.215 & 0.365 & \textbf{$+$20\%} \\
\rowcolor{oursrow}\quad \ours\trained & 0.606 & \textbf{0.389} & \textbf{0.272} & \textbf{0.622} & \textbf{0.316} & \textbf{0.227} & 0.446 & 0.248 & 0.152 & \textbf{0.562} & 0.321 & \textbf{0.219} & \textbf{0.368} & \textbf{$+$21\%} \\
\addlinespace[1pt]
\multicolumn{15}{@{}l}{\bfseries Qwen3-8B} \\[1pt]
\quad Zero-shot\frozen & 0.150 & 0.077 & 0.043 & 0.151 & 0.054 & 0.047 & 0.149 & 0.078 & 0.057 & 0.150 & 0.069 & 0.049 & 0.089 & \textbf{$-$73\%} \\
\quad Base harness\trained & 0.584 & 0.317 & 0.190 & 0.578 & 0.305 & 0.189 & 0.447 & 0.226 & 0.143 & 0.540 & 0.285 & 0.175 & 0.333 & --- \\
\quad DeepAgents\trained & 0.640 & 0.407 & 0.253 & 0.593 & 0.322 & 0.199 & 0.481 & 0.247 & 0.138 & 0.575 & 0.329 & 0.199 & 0.368 & \textbf{$+$10\%} \\
\quad INSPO\trained & 0.600 & 0.270 & 0.122 & 0.642 & 0.291 & 0.140 & 0.409 & 0.211 & 0.102 & 0.556 & 0.259 & 0.122 & 0.313 & \textbf{$-$6\%} \\
\quad EvolveR\trained & 0.610 & 0.347 & 0.187 & 0.599 & 0.314 & 0.209 & 0.471 & 0.219 & 0.140 & 0.564 & 0.296 & 0.180 & 0.347 & \textbf{$+$4\%} \\
\quad HarnessForge\trained & 0.540 & 0.310 & 0.200 & 0.583 & 0.282 & 0.180 & 0.307 & 0.157 & 0.113 & 0.483 & 0.254 & 0.167 & 0.301 & \textbf{$-$10\%} \\
\quad SkillRL\trained & 0.610 & 0.267 & 0.137 & 0.553 & 0.259 & 0.114 & \textbf{0.539} & 0.174 & 0.112 & 0.569 & 0.236 & 0.121 & 0.309 & \textbf{$-$7\%} \\
\rowcolor{oursrow}\quad + \Compression\trained & 0.627 & 0.340 & 0.210 & 0.613 & 0.274 & 0.181 & 0.495 & 0.286 & 0.188 & 0.582 & 0.301 & 0.193 & 0.359 & \textbf{$+$8\%} \\
\rowcolor{oursrow}\quad + \Steering\trained & \textbf{0.642} & \textbf{0.432} & \textbf{0.286} & 0.611 & 0.334 & 0.245 & 0.444 & 0.260 & 0.190 & 0.571 & 0.345 & 0.242 & 0.386 & \textbf{$+$16\%} \\
\rowcolor{oursrow}\quad \ours\trained & 0.616 & 0.423 & 0.283 & \textbf{0.659} & \textbf{0.394} & \textbf{0.259} & 0.520 & \textbf{0.313} & \textbf{0.221} & \textbf{0.602} & \textbf{0.379} & \textbf{0.256} & \textbf{0.412} & \textbf{$+$24\%} \\
\addlinespace[1pt]
\multicolumn{15}{@{}l}{\bfseries GPT-5-mini} \\[1pt]
\quad Zero-shot\frozen & 0.147 & 0.102 & 0.092 & 0.151 & 0.092 & 0.081 & 0.093 & 0.041 & 0.032 & 0.132 & 0.080 & 0.070 & 0.094 & --- \\
\midrule
\multicolumn{15}{@{}l}{\itshape Context-sufficient} \\[1pt]
\addlinespace[1pt]
\multicolumn{15}{@{}l}{\bfseries Qwen3-4B} \\[1pt]
\quad Zero-shot\frozen & 0.296 & 0.088 & 0.030 & 0.222 & 0.072 & 0.037 & 0.238 & 0.070 & 0.008 & 0.253 & 0.077 & 0.026 & 0.119 & \textbf{$-$67\%} \\
\quad Base harness\trained & 0.591 & 0.353 & 0.212 & 0.554 & 0.305 & 0.193 & 0.536 & 0.315 & 0.189 & 0.562 & 0.324 & 0.199 & 0.361 & --- \\
\quad DeepAgents\trained & 0.600 & 0.390 & 0.232 & 0.584 & 0.338 & 0.217 & 0.461 & 0.282 & 0.186 & 0.552 & 0.339 & 0.213 & 0.368 & \textbf{$+$2\%} \\
\quad INSPO\trained & 0.570 & 0.363 & \textbf{0.266} & 0.614 & 0.364 & \textbf{0.247} & 0.491 & 0.276 & 0.200 & 0.561 & 0.337 & 0.239 & 0.379 & \textbf{$+$5\%} \\
\quad EvolveR\trained & 0.640 & 0.380 & 0.197 & \textbf{0.683} & 0.344 & 0.220 & 0.524 & 0.276 & 0.168 & 0.619 & 0.336 & 0.196 & 0.384 & \textbf{$+$6\%} \\
\quad HarnessForge\trained & 0.597 & 0.373 & 0.252 & 0.635 & 0.346 & 0.200 & 0.524 & 0.323 & \textbf{0.228} & 0.588 & 0.348 & 0.226 & 0.388 & \textbf{$+$7\%} \\
\quad SkillRL\trained & 0.657 & 0.410 & 0.247 & 0.609 & 0.365 & 0.227 & 0.530 & 0.311 & 0.183 & 0.601 & 0.364 & 0.220 & 0.395 & \textbf{$+$9\%} \\
\rowcolor{oursrow}\quad + \Compression\trained & 0.623 & 0.323 & 0.177 & 0.607 & 0.316 & 0.200 & 0.539 & 0.283 & 0.192 & 0.592 & 0.309 & 0.190 & 0.363 & \textbf{$+$1\%} \\
\rowcolor{oursrow}\quad + \Steering\trained & 0.648 & 0.416 & 0.251 & 0.640 & 0.356 & 0.234 & 0.549 & 0.311 & 0.207 & 0.615 & 0.363 & 0.232 & 0.403 & \textbf{$+$12\%} \\
\rowcolor{oursrow}\quad \ours\trained & \textbf{0.680} & \textbf{0.429} & 0.251 & 0.663 & \textbf{0.390} & 0.245 & \textbf{0.560} & \textbf{0.357} & 0.226 & \textbf{0.637} & \textbf{0.393} & \textbf{0.241} & \textbf{0.424} & \textbf{$+$17\%} \\
\addlinespace[1pt]
\multicolumn{15}{@{}l}{\bfseries Qwen3-8B} \\[1pt]
\quad Zero-shot\frozen & 0.155 & 0.082 & 0.057 & 0.165 & 0.075 & 0.052 & 0.104 & 0.045 & 0.029 & 0.143 & 0.068 & 0.047 & 0.086 & \textbf{$-$79\%} \\
\quad Base harness\trained & 0.654 & 0.383 & 0.257 & 0.640 & 0.373 & 0.245 & 0.529 & 0.305 & 0.221 & 0.611 & 0.356 & 0.242 & 0.403 & --- \\
\quad DeepAgents\trained & 0.650 & 0.437 & 0.310 & 0.680 & 0.401 & 0.289 & 0.526 & 0.339 & 0.218 & 0.623 & 0.394 & 0.274 & 0.430 & \textbf{$+$7\%} \\
\quad INSPO\trained & 0.640 & 0.477 & \textbf{0.313} & 0.649 & 0.420 & 0.254 & 0.539 & 0.301 & 0.202 & 0.612 & 0.403 & 0.259 & 0.425 & \textbf{$+$5\%} \\
\quad EvolveR\trained & 0.617 & 0.430 & 0.282 & 0.643 & 0.416 & 0.292 & 0.503 & 0.328 & 0.228 & 0.591 & 0.394 & 0.269 & 0.418 & \textbf{$+$4\%} \\
\quad HarnessForge\trained & 0.620 & 0.417 & 0.253 & 0.592 & 0.356 & 0.239 & 0.413 & 0.240 & 0.170 & 0.547 & 0.341 & 0.223 & 0.370 & \textbf{$-$8\%} \\
\quad SkillRL\trained & \textbf{0.685} & 0.473 & 0.277 & 0.654 & 0.387 & 0.244 & 0.533 & 0.327 & 0.227 & 0.628 & 0.399 & 0.250 & 0.426 & \textbf{$+$6\%} \\
\rowcolor{oursrow}\quad + \Compression\trained & 0.645 & 0.436 & 0.279 & 0.685 & 0.418 & 0.279 & 0.516 & 0.277 & 0.193 & 0.619 & 0.381 & 0.253 & 0.418 & \textbf{$+$4\%} \\
\rowcolor{oursrow}\quad + \Steering\trained & 0.677 & 0.461 & 0.312 & \textbf{0.694} & \textbf{0.456} & \textbf{0.312} & 0.567 & 0.350 & 0.238 & 0.649 & 0.425 & 0.289 & 0.455 & \textbf{$+$13\%} \\
\rowcolor{oursrow}\quad \ours\trained & 0.684 & \textbf{0.479} & 0.312 & 0.686 & 0.433 & 0.300 & \textbf{0.583} & \textbf{0.375} & \textbf{0.269} & \textbf{0.654} & \textbf{0.431} & \textbf{0.295} & \textbf{0.460} & \textbf{$+$14\%} \\
\addlinespace[1pt]
\multicolumn{15}{@{}l}{\bfseries GPT-5-mini} \\[1pt]
\quad Zero-shot\frozen & 0.615 & 0.323 & 0.273 & 0.615 & 0.295 & 0.257 & 0.469 & 0.229 & 0.197 & 0.570 & 0.284 & 0.244 & 0.366 & --- \\
\bottomrule
\end{tabular*}
\caption{\textbf{CodeSearch validation, localization F1 by granularity.} The
per-granularity view of \cref{tab:cs_main}. The agent names a location at three
granularities and is scored at each:
\textbf{F}ile, \textbf{M}odule, the right class or top-level function inside it,
and \textbf{E}ntity, the right method. \emph{Overall} averages the three parts of
the \emph{Overall} split, so it stays an F1 on $[0,1]$. $\Delta$ is the relative
change in that average over the base harness of the same model, and bold is best within
a model. \trained\ marks an RL-trained model and \frozen\ one that was not fine-tuned.}
\label{tab:cs_gran}
\end{table*}

\subsection[Does the judge know when Steering is needed?]{Does the judge know when \Steering\ is needed?}
\label{app:gating}

\begin{figure*}[t]
\centering
\includegraphics[width=\textwidth]{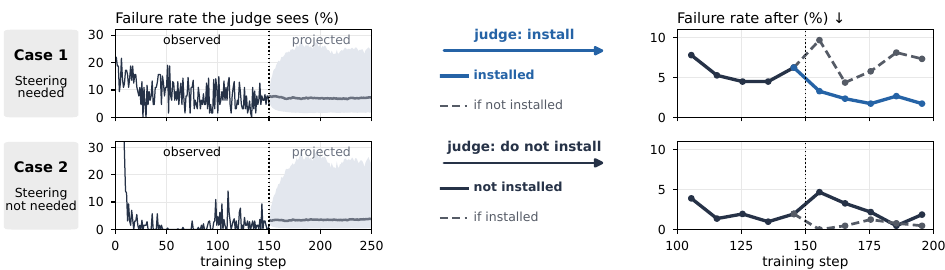}
\caption{\textbf{The judge's decision and the failure rate after it.} A failure is an
episode that ends without the model submitting an answer. The judge sees only the training
history before its decision. The runs with and without the guidance continue from the same
checkpoint, with two runs each in Case~1 and one each in Case~2.}
\label{fig:gating}
\end{figure*}

We build a setup to test a single \Steering\ program and its failure rate. Partway
through CodeSearch training, the judge reads how often one failure has occurred so far and
decides whether to install guidance for it (\cref{sec:steering}). We then continue training
twice from that point, once with the guidance and once without, so that we see the failure
rate after both decisions (\cref{fig:gating}). In Case~1, the failure keeps recurring, and the
judge installs the guidance. Installing it cuts the failure. In Case~2 the failure is already
rare, and the judge holds off. Without the guidance, the failure still falls as training continues. Installing it would also cut the failure, but it would add guidance for a behavior the policy learns on its own.

Case~1 of \cref{fig:gating} trains Qwen3-4B without \Compression, and Case~2 trains
Qwen3-8B with it.

\section{Harness Program Schema and Examples}
\label{app:programs}

\begin{figure*}[tp]
	\centering
	\includegraphics[width=\linewidth]{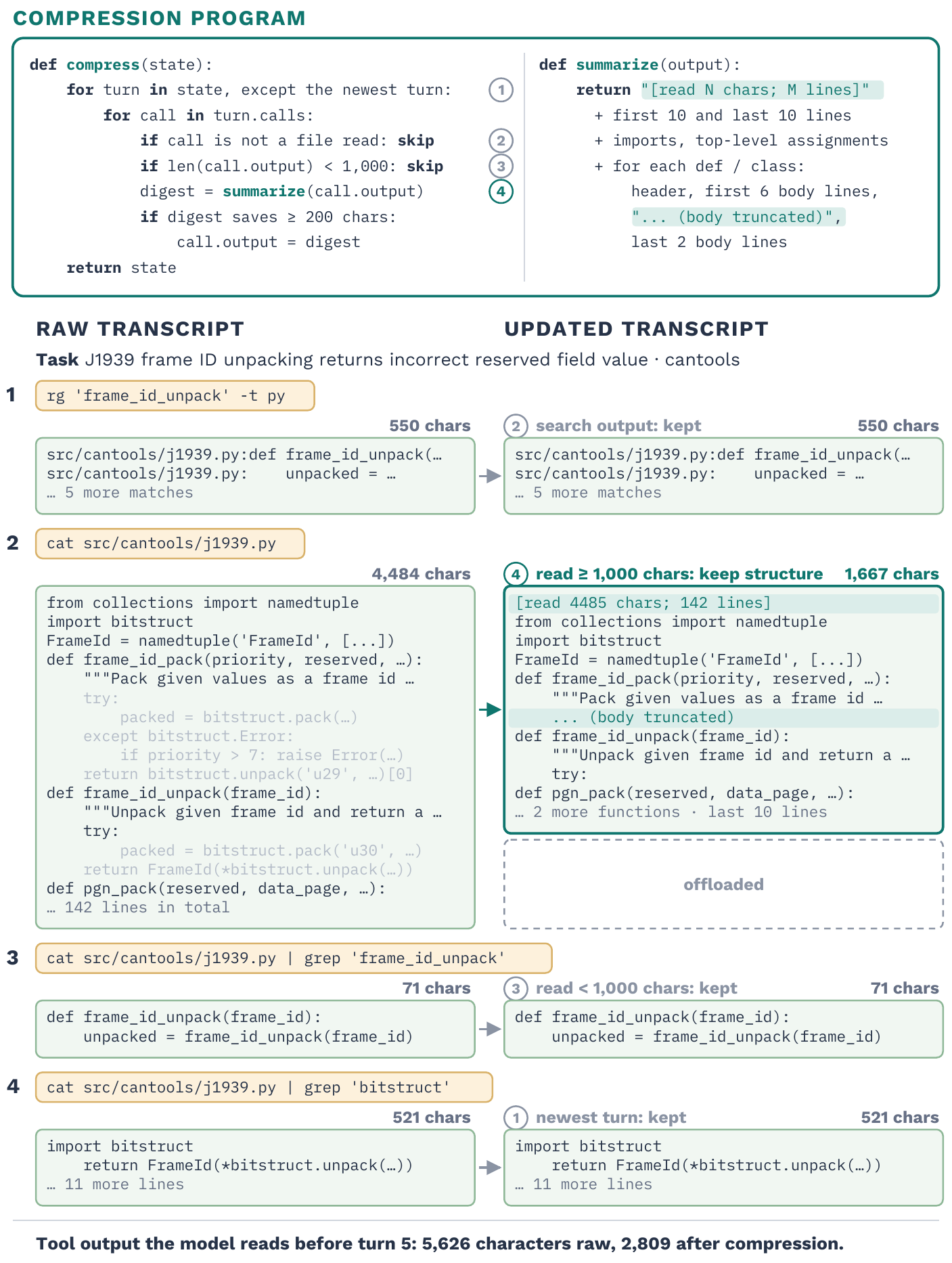}
	\caption{\textbf{A \compress\ program applied to a stored training episode.} Top: pseudocode
	of the generated program. Bottom: turns 1--4 of the episode as recorded (left) and as shown to
	the model (right); numbered rules match the program. Grey lines are dropped, and teal lines are
	written by the program.}
	\label{fig:compress_example}
\end{figure*}

\begin{figure*}[t]
	\centering
	\includegraphics[width=\linewidth]{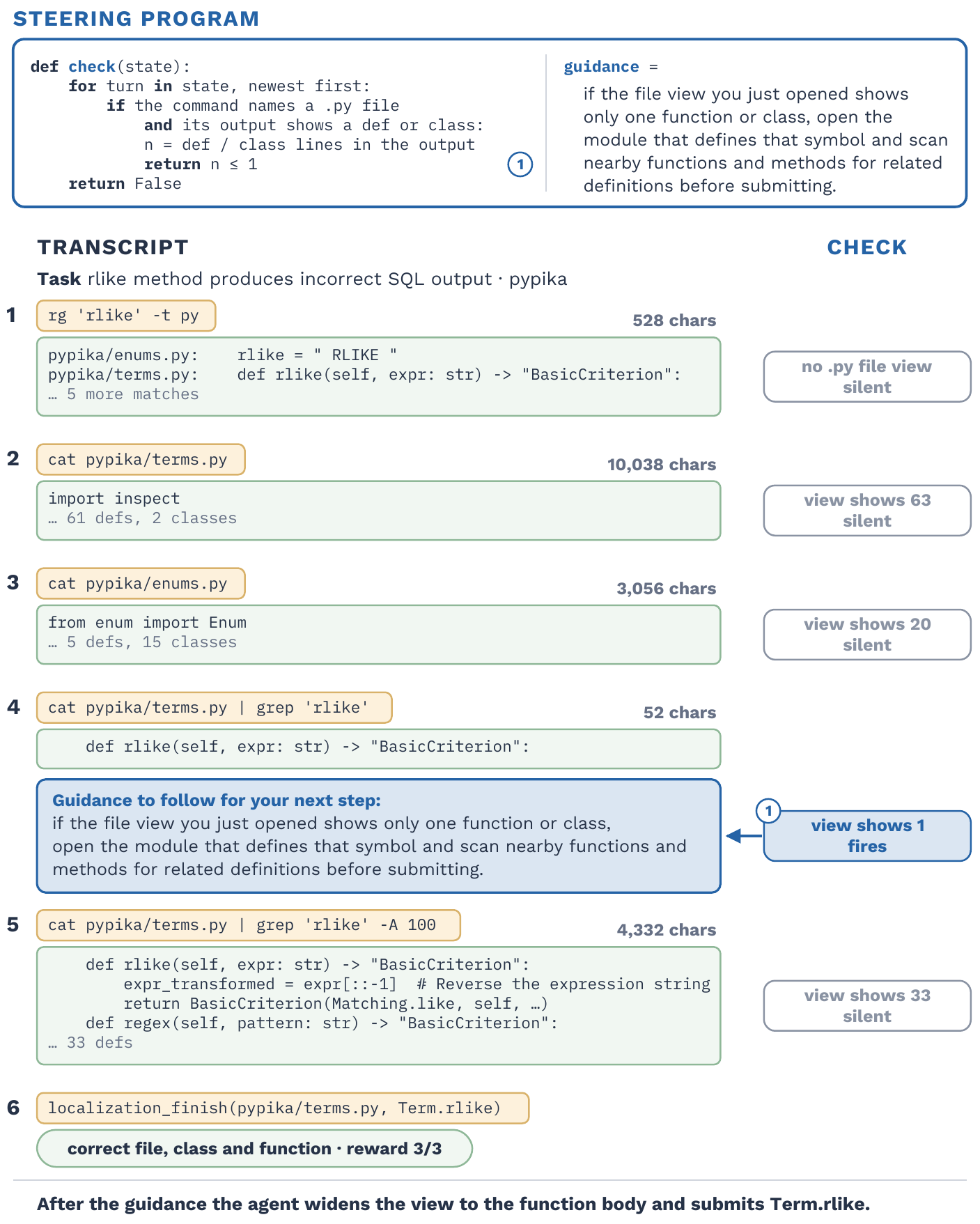}
	\caption{\textbf{A \steer\ trigger serving guidance in a live training episode.} Top: pseudocode
	of the generated check and its guidance. Bottom: the episode as recorded, with the check
	evaluated after every turn (right); the numbered rule matches the program. The check fires once,
	after turn 4, and the guidance is appended as a message before the model's next turn.}
	\label{fig:steer_example}
\end{figure*}

\Cref{fig:compress_example} and \cref{fig:steer_example} show an example \compress\ program and an example \steer\ trigger, each applied to a training episode.

\pp{\Compression\ example}
\Cref{fig:compress_example} applies a generated \compress\ program to a CodeSearch episode.
The program shortens only file reads of at least 1,000 characters in earlier turns, replacing
each with a digest that keeps the imports, the definitions, and the first and last lines.
Search output, short reads and the newest turn are kept as they are. In this episode, the tool
output the model reads before turn 5 falls from 5,626 to 2,809 characters, while the structure
of the file stays visible.

\pp{\Steering\ example}
\Cref{fig:steer_example} shows a generated \steer\ trigger in a CodeSearch training episode.
The check stays silent while the agent searches and opens whole files. After turn 4, the
agent's view shows a single function, so the check fires and the guidance asks the agent to
scan nearby definitions before submitting. The agent then widens the view to the function body
and submits \texttt{Term.rlike}, which earns the full reward of 3/3 for the correct file, class
and function.

\clearpage
\section{Prompt Details}
\label{app:prompts}

Each box below is simplified for space. \slot{\dots} marks text filled in at run time.

\begin{promptbox}{\textsc{ProposeCompress}: system prompt (simplified)}{compress}
You write compression programs for the harness of an LLM agent trained with reinforcement learning.

\slot{environment description}

Your program, \texttt{compress(view)}, runs before every turn. It gets the episode so far and returns it with earlier turns rewritten, and the agent is shown that instead. \slot{history budget}

A good program removes what the agent will not need and keeps what it will. A program fails in one of two ways:

\begin{itemize}[leftmargin=1.2em,itemsep=0pt,topsep=1pt,parsep=0pt]
  \item Unfaithful: it cuts what the agent still needs.
  \item Inert: it does not shorten the history enough to keep what the agent needs.
\end{itemize}
What the agent needs shows where it fetched content again, and each version of your program is measured on that content.

\prompthead{What your program reads}
\slot{fields of the view}

\prompthead{The agent's tools}
\slot{tool list}

\prompthead{Rules}
\begin{itemize}[leftmargin=1.2em,itemsep=0pt,topsep=1pt,parsep=0pt]
  \item Balance the two failures for this environment and budget.
  \item Cut more from older turns than from the newest one.
  \item Define \texttt{def compress(view)}, with allowed imports only and within a time limit.
  \item A program that fails is skipped, and the agent is shown its history uncompressed.
\end{itemize}
\end{promptbox}

\begin{promptbox}{\textsc{ProposeCompress}: writing programs (simplified)}{compress}
\prompthead{The episodes}
The agent played \slot{N} training tasks, once as served and once with its whole history. \slot{reward range, how often the context filled up, how often content was fetched again}

\prompthead{\slot{k} of these tasks, played both ways}
\slot{turn-by-turn episodes}

\prompthead{Your task}
Write \slot{n} programs, each from a different idea, each with a short intuition drawn from the episodes. Return the name, intuition and code as JSON.

\end{promptbox}

\begin{promptbox}{\textsc{ProposeCompress}: reviewing a program (simplified)}{compress}
\prompthead{The program}
\slot{name, intuition, code}

\prompthead{What it did to the episodes}
\slot{what the agent is shown with the program and without it}

\prompthead{The tasks it was written from}
\slot{episodes turn by turn, and the agent's last view with and without the program}

\prompthead{Your task}
This is review \slot{i} of \slot{n}. Accept the program if it removes what the agent does not need and keeps what it needs. Otherwise revise it, keeping its idea. Later reviews change less. Return a verdict, a short reason and any revised code as JSON.

\end{promptbox}

\begin{promptbox}{\textsc{ProposeCompress}: repairing a program (simplified)}{compress}
This compression program cannot be loaded: \slot{error}

\slot{code}

Fix it, and return the fixed code as JSON.

\end{promptbox}

\begin{promptbox}{\textsc{ProposeSteer}: system prompt (simplified)}{steer}
You write steering for the harness of an LLM agent trained with reinforcement learning.

\slot{environment description} \slot{reward description}

Steering shows the agent short guidance before the turns where a condition holds. You write it as a failure hypothesis with four parts: a failure check that says whether a finished episode made the failure, a trigger that decides before each turn whether to show the guidance, the guidance, and the insight it is written from. The trigger acts early, while the failure check judges at the end.

\prompthead{The failure check}
\begin{itemize}[leftmargin=1.2em,itemsep=0pt,topsep=1pt,parsep=0pt]
  \item Name a behaviour, not a low reward.
  \item Ground it in the turns.
  \item It may use the reward to show that the behaviour cost something.
  \item It may be an action the agent never took.
\end{itemize}
\prompthead{The trigger}
\begin{itemize}[leftmargin=1.2em,itemsep=0pt,topsep=1pt,parsep=0pt]
  \item Read only the episode so far.
  \item Fire on some episodes that go well if needed, but not on most.
  \item Fire on a mark in the turns, or on a phase of the episode.
  \item Fire while turns are left.
\end{itemize}
\prompthead{The guidance}
\begin{itemize}[leftmargin=1.2em,itemsep=0pt,topsep=1pt,parsep=0pt]
  \item Write it as ``If \slot{case}, \slot{action}.'', in terms the agent can see.
  \item Keep it general, with no names from the episodes.
  \item Keep it short and self-contained.
\end{itemize}
\prompthead{Inputs}
\slot{fields the trigger reads} \slot{the finished episode} \slot{tools}

\prompthead{Rules}
Write the two programs as Python functions that return \texttt{True} or \texttt{False}, using only allowed imports and within a time limit.

\end{promptbox}

\begin{promptbox}{\textsc{ProposeSteer}: one group of episodes (simplified)}{steer}
\prompthead{The task}
\slot{task statement} \slot{what the task wanted}

\prompthead{Its episodes}
\slot{the group's episodes at training step $k$, and how each ended}

\prompthead{Steering already installed}
\slot{installed hypotheses}

\prompthead{Your task}
Write one failure hypothesis from these episodes, or skip them. A weak hypothesis costs more than none. Return the name, insight, failure check, trigger and guidance as JSON, or a reason to skip.

\end{promptbox}

\begin{promptbox}{\textsc{ProposeSteer}: merging hypotheses (simplified)}{steer}
\prompthead{This round's failure hypotheses}
\slot{each hypothesis, with how often it fires and fails over the stored episodes}

\prompthead{Steering already installed}
\slot{installed hypotheses}

\prompthead{Your task}
Group the hypotheses by the failure they name, and write one hypothesis for each group worth steering, at most \slot{m}.

\begin{itemize}[leftmargin=1.2em,itemsep=0pt,topsep=1pt,parsep=0pt]
  \item Group those whose failure checks hold on the same episodes.
  \item Write each from what its members share.
  \item Weigh the replay: a check that holds on almost every episode, or a trigger that fires on many episodes that did not fail, is too broad.
  \item Do not repeat what the installed steering already covers.
\end{itemize}
Return the merged hypotheses as JSON.

\end{promptbox}

\begin{promptbox}{\textsc{ProposeSteer}: reviewing a hypothesis (simplified)}{steer}
\prompthead{The failure hypothesis}
\slot{name, insight, failure check, trigger, guidance}

\prompthead{What it does on the stored episodes}
\slot{how often the failure check holds, and how often the trigger fires on episodes that failed and on those that did not}

\prompthead{Example episodes}
\slot{examples of each kind, marking the turns where the guidance would be shown}

\prompthead{Your task}
This is review \slot{i} of \slot{n}. Check two things.

\begin{itemize}[leftmargin=1.2em,itemsep=0pt,topsep=1pt,parsep=0pt]
  \item The failure check holds on the episodes that made the failure, and on no others.
  \item The trigger reaches those episodes while turns are left, and fires on few that go well.
\end{itemize}
A hypothesis that fires or holds too broadly is not offered. Accept it, or revise it keeping its idea. Later reviews change less. Return a verdict, a short reason and any revision as JSON.

\end{promptbox}

\begin{promptbox}{\textsc{Admit}: system prompt (simplified)}{steer}
You judge whether to steer an LLM agent that is trained with reinforcement learning away from a failure.

\slot{environment description} \slot{reward description}

Guidance costs room in the agent's context and can stand in the way of what the agent would learn on its own. So it is worth showing only for a failure the policy does not fix by training. You are shown one failure hypothesis and its failure curve: the share of each training step's episodes that made the failure, with a fitted line and a 95\% band.

Return one of three verdicts:

\begin{itemize}[leftmargin=1.2em,itemsep=0pt,topsep=1pt,parsep=0pt]
  \item intervene: the failure holds steady or rises. Its trigger and guidance go to the gate.
  \item monitor: the failure looks real but has not settled. It is judged again at the next optimization.
  \item continue: the failure keeps falling. It is left to training.
\end{itemize}
\end{promptbox}

\end{document}